\documentclass[10pt,twocolumn,letterpaper]{article}
\usepackage[T1]{fontenc}

\usepackage[pagenumbers]{cvpr}
\usepackage{float}
\makeatletter
\let\kernel@addtocurcol\@addtocurcol
\usepackage{flafter}
\let\flafter@addtocurcol\@addtocurcol
\let\@addtocurcol\kernel@addtocurcol
\makeatother

\usepackage{booktabs}
\usepackage{amsmath}
\usepackage{amssymb}
\usepackage{xspace}
\newcommand{\ours}{FastWild-GS\xspace}   %

\newcommand{\cOursBestPSNR}{25.52}
\newcommand{\cOursBestSSIM}{0.874}

\newcommand{\cLitBestPSNR}{25.62}

\newcommand{\cOursMin}{8.8}

\newcommand{\cMinSceneMin}{6.1} %

\newcommand{\cAsymRatio}{24} %

\newcommand{\cAsymH}{3.50}

\definecolor{cvprblue}{rgb}{0.21,0.49,0.74}
\usepackage[pagebackref,breaklinks,colorlinks,allcolors=cvprblue]{hyperref}
\newcommand{\extdef}[2]{\expandafter\gdef\csname ext:#1\endcsname{#2}}
\DeclareRobustCommand{\extref}[1]{%
  \ifcsname ext:#1\endcsname\csname ext:#1\endcsname
  \else\textbf{??}\PackageWarning{xrefs}{`#1' is not listed in xrefs.tex}\fi}
\extdef{tab:supp_time}{Tab.~S2}
\extdef{tab:supp_hparams}{Tab.~S1}
\extdef{tab:supp_size}{Tab.~S3}
\extdef{tab:supp_ablation_scene}{Tab.~S4}
\extdef{tab:supp_osr}{Tab.~S5}
\extdef{tab:supp_std}{Tab.~S7}
\extdef{tab:supp_std_bb}{Tab.~S8}
\extdef{fig:supp_errmap}{Fig.~S1}
\extdef{sec:supp_impl}{Sec.~A}
\extdef{sec:supp_analysis}{Sec.~C}
\extdef{tab:main}{Tab.~1}
\extdef{fig:teaser}{Fig.~2}
\extdef{fig:zoom}{Fig.~5}
\extdef{fig:osr}{Fig.~8}
\extdef{tab:osr}{Tab.~2}
\extdef{tab:ablation}{Tab.~3}
\extdef{tab:tta_suppl}{Tab.~4}
\extdef{tab:backbones}{Tab.~5}
\extdef{eq:radial}{Eq.~(5)}
\extdef{sec:scoring}{Sec.~3.3}

\def\confName{CVPR}
\def\confYear{2027}

\title{FastWild-GS: Minutes-Scale In-the-Wild Gaussian Splatting}

\author{Donghun Ryu$^{*}$ \qquad Seokhyun Kim$^{*}$ \qquad Minhyeok Lee\\
Chung-Ang University\\
{\tt\small \{donghun0621, lll6610, mlee\}@cau.ac.kr}}

\begin{document}
\twocolumn[{%
\renewcommand\twocolumn[1][]{#1}%
\maketitle
\begin{center}
\centering
\includegraphics[width=\textwidth]{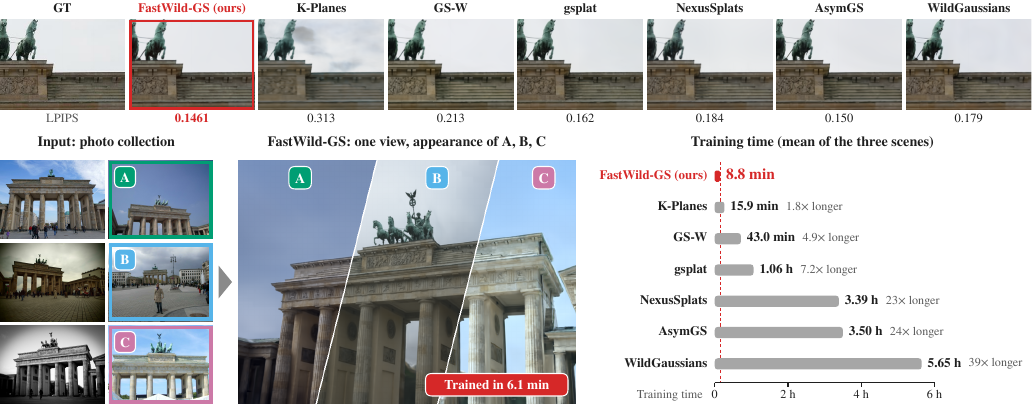}
\captionof{figure}{Top: a crop of one test view rendered by \ours and recent methods, with the mean LPIPS on PhotoTourism.
Bottom: from a photo collection, \ours renders one view with the appearance of three of its images (A, B, C) after \cMinSceneMin\ minutes of training on this scene; the training times on the right are means of the three scenes.}
\label{fig:first}
\end{center}%
}]
{\renewcommand{\thefootnote}{\fnsymbol{footnote}}\footnotetext[1]{Equal contribution.}}
\begin{abstract}
Reconstructing a 3D scene from casual Internet photographs, a setting known as in-the-wild reconstruction, is challenging because the images are captured by different people and cameras at different times, producing large variations in exposure, illumination and white balance.
Existing high-quality approaches mostly require hours of per-scene optimization.
3D Gaussian Splatting (3DGS) methods for controlled captures train a scene in minutes, but on photo collections they average the appearance of all images, and their densification and pruning count color differences between images as reconstruction error.
We propose \ours, which adapts both the colors and the density control of a Gaussian scene to each image.
Its appearance module predicts the color of every Gaussian for each image, and color-matched scoring applies the per-image color transform of this module to the renders that score Gaussians, which compensates for exposure and white-balance differences in the error that drives densification and pruning.
On PhotoTourism, \ours trains $\cAsymRatio\times$ faster than AsymGS, the strongest baseline in PSNR, with a higher SSIM at nearly the same PSNR (\cOursBestPSNR\ against \cLitBestPSNR\,dB), and on NeRF-OSR it reaches the best mean PSNR and SSIM in 9.7 minutes per scene.
With the appearance module, six other 3DGS backbones reach 24.6 to 25.1\,dB PSNR on PhotoTourism.
\end{abstract}
\section{Introduction}
\label{sec:intro}

\begin{figure*}[t]
\centering
\includegraphics[width=\textwidth]{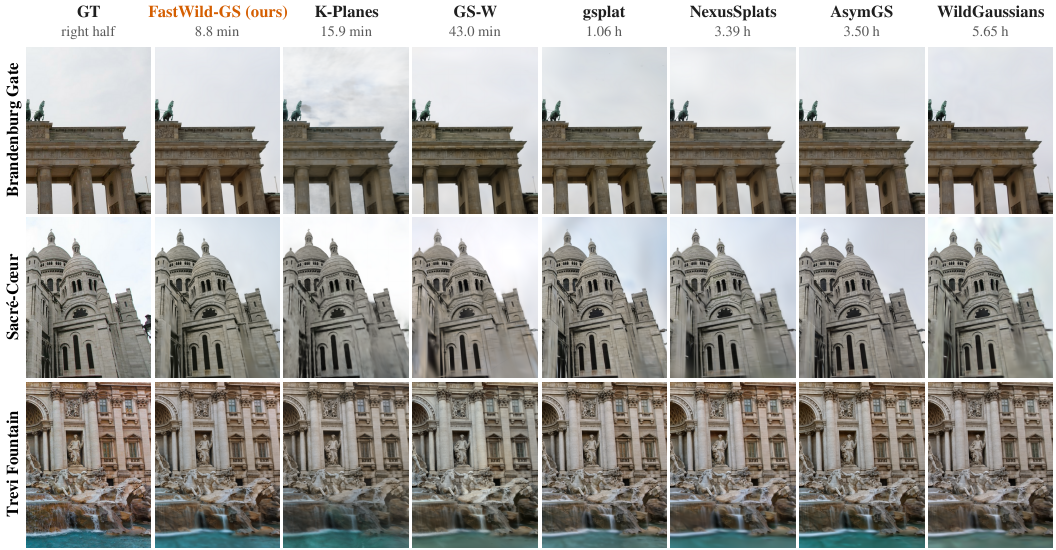}
\caption{Right halves of PhotoTourism test views rendered by \ours and recent methods, ordered by training time.
\ours trains in minutes, while the other methods train for up to several hours.}
\label{fig:teaser}
\end{figure*}

Novel-view synthesis from casual Internet photo collections~\cite{snavely2006photo}, the in-the-wild setting, is more difficult than reconstruction from a controlled capture.
The photographs are taken by different users and cameras at different times of day and in different seasons and weather conditions.
Consequently, the same physical surface appears with different brightness, color or illumination across images.

Standard 3D Gaussian Splatting (3DGS)~\cite{kerbl2023gaussian} assumes that a shared scene representation explains the appearance of all input images, and averages these differences.
Existing in-the-wild methods introduce image-specific appearance models~\cite{martinbrualla2021nerfw,kulhanek2024wildgaussians}, but recent methods mostly take hours of per-scene optimization (\cref{tab:main}).

3DGS methods for controlled captures train a scene in minutes~\cite{ren2026fastgs,chen2025dashgaussian,hanson2025speedy}, but on photo collections, a color shared by all images cannot reproduce the images (\cref{tab:backbones}), and density control, the densification and pruning of Gaussians according to the rendering error~\cite{ren2026fastgs}, also counts the color differences between images as reconstruction error, so an appearance module does not by itself restore the rendering quality (\cref{tab:ablation}).
\ours keeps the training in minutes and adapts both the colors and the density control of a 3DGS backbone to each image.
Its appearance module predicts the color of each Gaussian from a vector learned for each image, the Photo Appearance Token (PAT), and a channel transform adjusts the exposure and white balance of each render.
Color-matched scoring applies the same transform to the renders from which density control measures the error.
Replacing the appearance module, including its channel transform, with that of WildGaussians~\cite{kulhanek2024wildgaussians} lowers PSNR by 1.98\,dB and lengthens the training from \cOursMin\ to 12.4 minutes (\cref{tab:ablation}).
\ours trains an in-the-wild scene in \cOursMin\ minutes at the rendering quality of methods trained for hours (\cref{tab:main,fig:first,fig:teaser}).
Our contributions are:
\begin{enumerate}
\item \textbf{Per-image appearance module and color-matched scoring} that adapt a 3DGS backbone to photo collections: the colors are predicted for each image, and density control scores Gaussians on color-matched renders.
\item \textbf{Minutes-scale training} that is $\cAsymRatio\times$ faster than AsymGS with a higher SSIM on PhotoTourism and gives the best mean PSNR and SSIM on NeRF-OSR.
\item \textbf{General applicability}: the appearance module extends six other 3DGS backbones to photo collections.
\end{enumerate}
\section{Related Work}
\label{sec:related}

\paragraph{3D Gaussian Splatting and training acceleration.}
Neural radiance fields~\cite{mildenhall2021nerf} established optimization-based novel-view synthesis from posed photographs, and multiresolution hash encoding~\cite{muller2022instantngp} reduced their training cost.
3DGS~\cite{kerbl2023gaussian} represents a scene with optimized anisotropic Gaussian primitives and enables fast rasterized rendering.
FastGS~\cite{ren2026fastgs} accelerates 3DGS with multi-view consistent densification and pruning.
DashGaussian~\cite{chen2025dashgaussian} schedules rendering resolution together with the primitive count so that the count stays low early in training, as our growth schedule does (\cref{sec:growth}).
Speedy-Splat~\cite{hanson2025speedy} prunes aggressively right before opacity resets and at fixed intervals after densification.
Other methods schedule the Gaussian count or adapt splitting~\cite{dhiman2026turbo,edc2024,adpsplit2026,lyu2026sadgs}.
These methods target photometrically consistent captures; color-matched scoring (\cref{sec:scoring}) addresses density control on photo collections.

\paragraph{In-the-wild appearance modeling.}
NeRF-W~\cite{martinbrualla2021nerfw} introduced the standard PhotoTourism setting and per-image appearance embeddings.
Subsequent NeRF-based methods refine the same setting with hallucinated appearance transfer, cross-ray feature aggregation and multi-sequence modeling~\cite{chen2022hanerf,yang2023crnerf,li2023nerfms}.
Gaussian Splatting methods such as Wild-GS~\cite{xu2024wildgs}, GS-W~\cite{zhang2024gsw}, WE-GS~\cite{wang2024wegs}, Splatfacto-W~\cite{nerfstudio2023splatfactow} and WildGaussians~\cite{kulhanek2024wildgaussians} use image-conditioned, transformation-based or multi-scale mechanisms to model appearance variation, and WilLaGS~\cite{willags} conditions a tri-plane appearance field on a per-image latent code.
Hierarchical 3DGS~\cite{kerbl2024hierarchical} corrects exposure with a per-image affine color transform.

\paragraph{Robust and efficient in-the-wild 3DGS variants.}
Recent work focuses on appearance modeling or distractor handling~\cite{dahmani2024swag,nexussplats,sabour2024spotlesssplats,asymgs,forestsplats,fu2025robustsplat,robustsplatpp2025,lin2025davigs,li2025mwgs}, and other variants address efficiency~\cite{wang2025wlgs,wang2025lookatthesky,chen2026threedgsw}.
RobustSplat++~\cite{robustsplatpp2025} delays Gaussian growth to keep densification from modeling transients and illumination changes.
Most of these in-the-wild methods take hours of per-scene optimization (\cref{tab:main}).
In contrast, \ours trains in minutes with an appearance module that follows the per-image code and per-Gaussian feature conditioning of WildGaussians, and applies its channel transform to the scoring renders instead of delaying Gaussian growth.
\section{Method}
\label{sec:method}

\begin{figure*}[t]
\centering
\includegraphics[width=\textwidth]{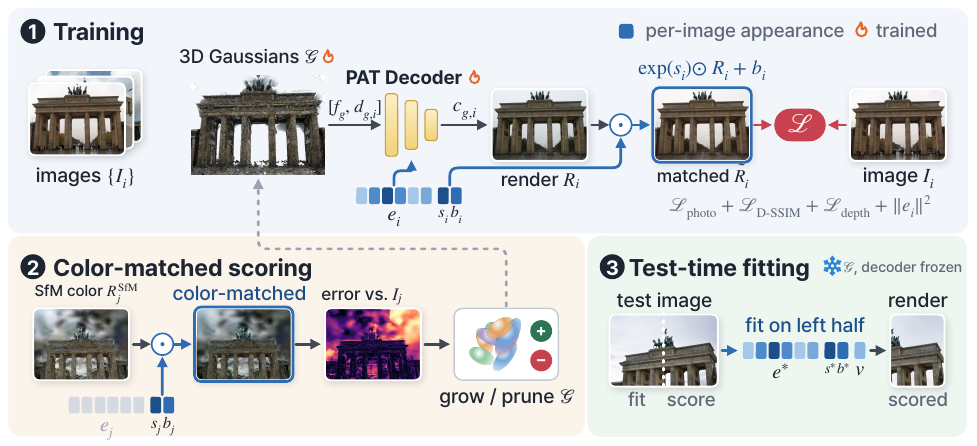}
\caption{\textbf{Overview of \ours.}
(1) Training: the PAT Decoder predicts the color $c_{g,i}$ of every Gaussian from its Splat Descriptor $[f_g, d_{g,i}]$ and the PAT $e_i$ of image $I_i$, and the channel transform $\exp(s_i)\odot R_i+b_i$ is applied to the render before the loss; a monocular depth prior regularizes the rendered depth.
(2) Color-matched scoring: the same channel transform is applied to the renders that score Gaussians for densification and pruning.
(3) Test-time fitting: the Gaussians and the decoder are frozen; a test-time PAT $e^\ast$, the transform $(s^\ast,b^\ast)$ and a radial gain $v$ are fitted on the left half of each test image, and the right half is scored.}
\label{fig:overview}
\end{figure*}

\ours adds to a shared Gaussian scene an appearance module that consists of a PAT per image, a PAT Decoder shared across the scene and a per-image channel transform.
It applies the same transform to the renders that score Gaussians, and fits the PAT, the transform and a radial gain to each test view (\cref{fig:overview}).

\subsection{Preliminaries}
\label{sec:backbone}
The scene is a set of 3D Gaussians initialized from the SfM points of the scene, each with a position, scale, rotation and opacity, as in 3DGS~\cite{kerbl2023gaussian}.
We adopt the rasterizer and the multi-view consistent densification and pruning of FastGS~\cite{ren2026fastgs}: at each densification and pruning step, a set of training views is rendered, and each Gaussian receives a densification score and a pruning score from the error between these renders and the images; a high score selects the Gaussian for densification or for pruning.
The densification thresholds are fixed until densification ends.

\subsection{Appearance Module}
\label{sec:appearance}
\begin{figure*}[t]
\centering
\includegraphics[width=\textwidth]{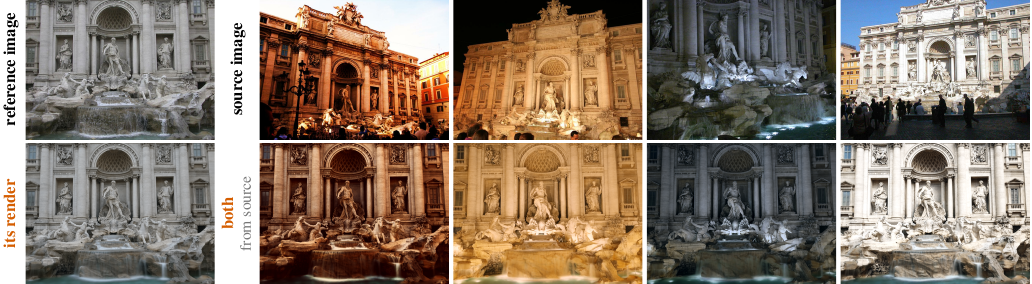}
\caption{Appearance module on a fixed view.
Bottom: the reference view rendered with the PAT and the channel transform of the source image above.}
\label{fig:appearance}
\end{figure*}

In 3DGS, the color of a Gaussian depends only on the view direction, but the images of a photo collection show the same surface in different colors.
The appearance module predicts the color of each Gaussian for each image: a PAT Decoder shared across the scene takes the PAT of the image and the Splat Descriptor of the Gaussian, and a per-image channel transform is applied after rasterization (\cref{fig:appearance}).
The PAT and channel transform are 134 parameters per image.

\paragraph{Photo Appearance Token.}
Each training image $I_i$ has a learnable PAT $e_i\in\mathbb{R}^{128}$, a per-image appearance embedding~\cite{martinbrualla2021nerfw} and the only per-image input of the PAT Decoder.

\paragraph{Splat Descriptor and PAT Decoder.}
Each Gaussian has a feature $f_g\in\mathbb{R}^{32}$; with the view direction $d_{g,i}$ it forms the Splat Descriptor $[f_g, d_{g,i}]$.
The PAT Decoder, an MLP shared across the scene, takes the Splat Descriptor and the PAT and outputs the color of Gaussian $g$ in image $i$:
\begin{equation}
c_{g,i} = \sigma\big(\mathrm{MLP}([f_g, d_{g,i}, e_i])\big),
\label{eq:color}
\end{equation}
where $d_{g,i}$ is the unit direction from the camera of image $i$ to the Gaussian and $\sigma$ is the sigmoid.
Because the decoder produces the whole color, the Gaussians store no learned color, whereas WildGaussians~\cite{kulhanek2024wildgaussians} corrects a color stored in each Gaussian.
The PAT is a single vector per image, but because the decoder also receives the Splat Descriptor, the color change between images differs for each Gaussian.

\paragraph{Channel transform.}
Images of the same place also differ globally in exposure and white balance.
After rasterization, the render $R_i$ of image $i$ is matched to the image by
\begin{equation}
\tilde{R}_i = \exp(s_i)\odot R_i + b_i, \qquad s_i, b_i\in\mathbb{R}^3,
\label{eq:transform}
\end{equation}
with $s_i$ a channel-wise log-scale and $b_i$ a channel-wise bias.
The transform is applied outside the decoder, so the same transform can be applied to the scoring renders (\cref{sec:scoring}).

\begin{table*}[t]
\centering
\footnotesize
\setlength{\tabcolsep}{3pt}
\resizebox{\textwidth}{!}{%
\begin{tabular}{@{}l ccc ccc ccc ccc r@{}}
\toprule
 & \multicolumn{3}{c}{Brandenburg Gate} & \multicolumn{3}{c}{Sacr\'e-C\oe ur} & \multicolumn{3}{c}{Trevi Fountain} & \multicolumn{3}{c}{Mean} & \\
\cmidrule(lr){2-4}\cmidrule(lr){5-7}\cmidrule(lr){8-10}\cmidrule(lr){11-13}
Method & PSNR$\uparrow$ & SSIM$\uparrow$ & LPIPS$\downarrow$ & PSNR$\uparrow$ & SSIM$\uparrow$ & LPIPS$\downarrow$ & PSNR$\uparrow$ & SSIM$\uparrow$ & LPIPS$\downarrow$ & PSNR$\uparrow$ & SSIM$\uparrow$ & LPIPS$\downarrow$ & Train \\
\midrule
K-Planes~\cite{fridovichkeil2023kplanes} & 23.65 & 0.859 & 0.253 & 19.96 & 0.762 & 0.299 & 19.70 & 0.662 & 0.388 & 21.10 & 0.761 & 0.313 & 15.9\,min \\
GS-W~\cite{zhang2024gsw} & 24.35 & 0.903 & 0.155 & 19.73 & 0.824 & 0.210 & 20.06 & 0.724 & 0.272 & 21.38 & 0.817 & 0.213 & 43.0\,min \\
Scaffold-GS~\cite{lu2024scaffold} & 25.45 & 0.923 & 0.127 & 21.85 & 0.871 & 0.157 & 23.21 & 0.768 & 0.228 & 23.50 & 0.854 & 0.170 & 45.9\,min \\
gsplat~\cite{ye2025gsplat} & 26.36 & 0.931 & 0.118 & 22.05 & 0.876 & 0.154 & 22.56 & 0.765 & 0.213 & 23.66 & 0.857 & 0.162 & 1.06\,h \\
WildGaussians~\cite{kulhanek2024wildgaussians} & 27.77 & 0.927 & 0.133 & 22.56 & 0.859 & 0.177 & 23.63 & 0.766 & 0.228 & 24.65 & 0.851 & 0.179 & 5.65\,h \\ %
NexusSplats~\cite{nexussplats} & 27.76 & 0.922 & 0.141 & 23.13 & 0.859 & 0.174 & 23.97 & 0.766 & 0.237 & 24.95 & 0.849 & 0.184 & 3.39\,h \\ %
AsymGS~\cite{asymgs} & 28.56 & 0.938 & \textbf{0.109} & 23.78 & 0.887 & 0.139 & \textbf{24.52} & 0.790 & 0.202 & \textbf{25.62} & 0.872 & 0.150 & 3.50\,h \\ %
ForestSplats~\cite{forestsplats} & \multicolumn{9}{c}{} & 25.36 & 0.871 & 0.151 & 7.9\,h$^\dagger$ \\ %
WilLaGS~\cite{willags} & 28.45 & 0.937 & 0.134 & \textbf{23.80} & 0.889 & 0.150 & 24.27 & \textbf{0.822} & \textbf{0.168} & 25.51 & \textbf{0.883} & 0.151 & 30.8\,min \\ %
\midrule
\ours & \textbf{28.87} & \textbf{0.940} & \textbf{0.1086} & 23.72 & \textbf{0.893} & \textbf{0.1357} & 23.96 & 0.790 & 0.1941 & 25.52 & 0.874 & \textbf{0.1461} & \textbf{8.8\,min} \\ %
\quad std over seeds & 0.39 & 0.0001 & 0.0029 & 0.20 & 0.0009 & 0.0017 & 0.04 & 0.0010 & 0.0021 & 0.15 & 0.0003 & 0.0017 & 0.07\,min \\
\bottomrule
\end{tabular}%
}
\caption{PhotoTourism per scene.
$^\dagger$: training time reported in the paper.
Bold: best per column.}
\label{tab:main}
\end{table*}

\begin{figure*}[t]
\centering
\includegraphics[width=\textwidth]{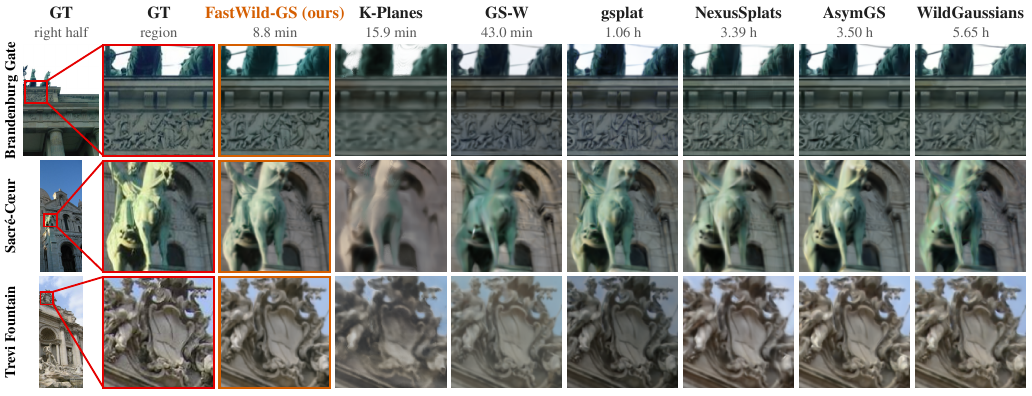}
\caption{Enlarged regions of further PhotoTourism test views.
The first column shows the right half of the test view with the enlarged region in red.}
\label{fig:zoom}
\end{figure*}

\subsection{Color-Matched Scoring}
\label{sec:scoring}
Densification and pruning should respond to errors that adding or removing Gaussians can fix.
The scores of FastGS come from separate scoring renders compared with the images, and an appearance model in the training loss does not affect them.
The Gaussians store no learned color, so the scoring renders use a color that does not depend on any image: each Gaussian keeps a fixed color inherited from the SfM points.
Their error therefore also contains the global color difference between each image and the scene, which no change of Gaussians can fix.
We apply the channel transform learned for each view,
\begin{equation}
\hat{R}_j = \exp(s_j)\odot R^{\text{SfM}}_j + b_j,
\label{eq:scoring}
\end{equation}
to its scoring render $R^{\text{SfM}}_j$ before the error is computed.
This reduces the global color difference, while the scoring color stays independent of the learned appearance.
Color-matched scoring adds no renders to those of FastGS.

\subsection{Growth Schedule}
\label{sec:growth}
\begin{figure}[t]
\centering
\includegraphics[width=\columnwidth]{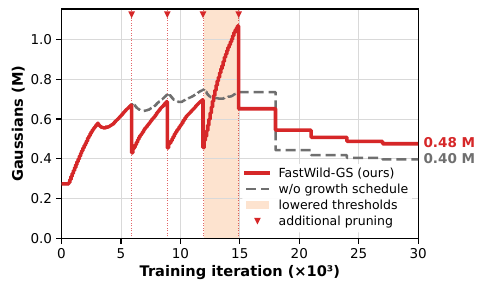}
\caption{Number of Gaussians during training with and without the growth schedule (mean of the three PhotoTourism scenes).}
\label{fig:growth}
\end{figure}

The cost of an iteration grows with the number of Gaussians~\cite{chen2025dashgaussian}, so a Gaussian added late in training adds less to the training time than one added early.
We lower the densification thresholds in the last iterations of densification, which adds Gaussians and raises the rendering quality without increasing the training time (\cref{fig:growth,tab:ablation}).
To offset the cost of the added Gaussians, we place additional pruning steps before the opacity resets, where Speedy-Splat~\cite{hanson2025speedy} prunes, and before densification ends.

\subsection{Training}
\label{sec:training}
The Gaussians, their features, the PATs, the PAT Decoder and the channel transforms are optimized jointly with
\begin{equation}
\begin{aligned}
\mathcal{L} &= 0.6\,\mathcal{L}_{\text{photo}}(\tilde{R}_i, I_i) + 0.4\,\big(1-\mathrm{SSIM}(\tilde{R}_i, I_i)\big)\\
&\quad + \lambda_{\text{d}}(t)\,\mathcal{L}_{\text{depth}} + 10^{-4}\lVert e_i\rVert_2^2,
\end{aligned}
\label{eq:loss}
\end{equation}
where $\mathcal{L}_{\text{photo}}$ is the mean of an L1 term and a root-mean-square term.
$\mathcal{L}_{\text{depth}}$ is a smooth-L1 loss between the logarithms of the rendered depth and of a monocular depth prior from Depth Anything 3~\cite{lin2026depthanything3}, aligned to the scale of the SfM points of the training image.
The prior constrains the geometry with a signal that does not depend on the appearance of the image; the schedule of its weight $\lambda_{\text{d}}(t)$ is in \extref{tab:supp_hparams}.
Parameters are averaged over the last 2k iterations.

\begin{figure*}[t]
\centering
\includegraphics[width=\textwidth]{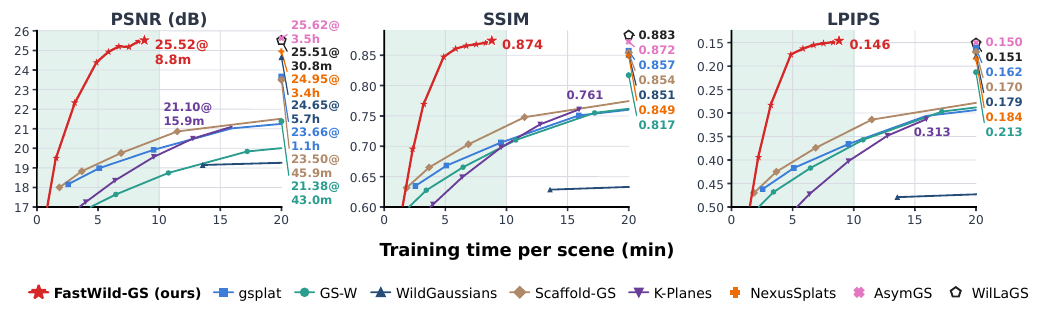}
\caption{Quality against training time per scene on PhotoTourism.
Methods that train longer than 20 minutes are shown at the right edge with their quality metric and training time.}
\label{fig:frontier}
\end{figure*}

\subsection{Test-Time Fitting}
\label{sec:tta}
Given a test view, we follow the half-image protocol of NeRF-W~\cite{martinbrualla2021nerfw}: the per-image appearance parameters are fitted on the left half and metrics are computed on the right half.
The Gaussians, their features and the PAT Decoder stay fixed.
We fit a test-time PAT $e^\ast$, the transform $(s^\ast, b^\ast)$ of \cref{eq:transform} and a radial gain $v\in\mathbb{R}^3$, 137 parameters in total.
The radial gain models a brightness fall-off from the image center, such as lens vignetting, which differs between cameras and which the channel transform, the same for all pixels, cannot represent:
\begin{equation}
\begin{aligned}
\tilde{R}^\ast &= \exp(s^\ast)\odot\big(G\odot R^\ast\big) + b^\ast,\\
G_c(x) &= \exp\big(\tanh(v_c)\,\bar\rho(x)\big),
\end{aligned}
\label{eq:radial}
\end{equation}
where $R^\ast$ is the render with $e^\ast$, $c$ indexes the color channel, and $\bar\rho(x)$ is the squared distance of pixel $x$ from the image center, in coordinates normalized to $[-1,1]$, minus its mean over the image.
\section{Experiments}
\label{sec:experiments}

\subsection{Setup}
\label{sec:setup}

\paragraph{Datasets and metrics.}
We evaluate on the three standard PhotoTourism scenes~\cite{snavely2006photo,martinbrualla2021nerfw}, Brandenburg Gate, Sacr\'e-C\oe ur and Trevi Fountain, and on four NeRF-OSR scenes~\cite{rudnev2022nerfosr}, Europa, LWP, ST and St.\ Johann, which are outdoor captures under varying illumination.
We report PSNR, SSIM~\cite{wang2004ssim} and LPIPS (AlexNet)~\cite{zhang2018lpips} under the half-image protocol of \cref{sec:tta}.

\paragraph{Implementation.}
\ours is trained for 30k iterations per scene with the same hyperparameters on both datasets (\extref{tab:supp_hparams}).
Our results are means over five seeds.

\paragraph{Baselines.}
The metrics of the compared methods are taken from their papers or from NerfBaselines~\cite{kulhanek2024nerfbaselines}.
Because the training times in the original papers come from different hardware, we measure all training times on a single RTX~5090 GPU, except those marked with~$\dagger$, which are taken from the original papers.
Methods with public code are trained with the code released by their authors or through NerfBaselines.
WilLaGS~\cite{willags}, NeRF-MS~\cite{li2023nerfms}, SWAG~\cite{dahmani2024swag} and WE-GS~\cite{wang2024wegs} have no public code, and we measure their training times by reproducing them from the descriptions in their papers.

\subsection{PhotoTourism}
\label{sec:results}

\ours achieves the shortest training time, \cOursMin\ minutes per scene, with \cOursBestPSNR\,dB PSNR and \cOursBestSSIM\ SSIM (\cref{tab:main,fig:frontier}).
AsymGS~\cite{asymgs}, the best in PSNR at \cLitBestPSNR\,dB, takes \cAsymH\ hours per scene, and \ours trains $\cAsymRatio\times$ faster with a higher SSIM and a lower LPIPS.
The half-image evaluation of WilLaGS~\cite{willags} gives AsymGS 25.03\,dB, 0.850 SSIM and 0.186 LPIPS, worse than \ours in all three metrics (\extref{sec:supp_analysis}).
K-Planes~\cite{fridovichkeil2023kplanes}, the next fastest at 15.9 minutes, is 4.42\,dB lower in PSNR, and WilLaGS~\cite{willags}, at 25.51\,dB and 0.883 SSIM, trains 3.5$\times$ longer.
\ours preserves the Brandenburg Gate frieze relief and the Trevi Fountain ornament at a sharpness close to that of the methods trained for hours (\cref{fig:zoom}).

\subsection{NeRF-OSR}
\label{sec:osr}

\begin{table}[t]
\centering
\small
\setlength{\tabcolsep}{3pt}
\begin{tabular}{@{}l ccc r@{}}
\toprule
Method & PSNR$\uparrow$ & SSIM$\uparrow$ & LPIPS$\downarrow$ & Train \\
\midrule
NeRF-MS~\cite{li2023nerfms} & 21.61 & 0.716 & 0.317 & 2.11\,h \\ %
SWAG~\cite{dahmani2024swag} & 23.00 & 0.801 & 0.270 & 44.3\,min \\ %
GS-W~\cite{zhang2024gsw} & 23.79 & 0.792 & 0.217 & 1.77\,h \\ %
WE-GS~\cite{wang2024wegs} & 23.91 & 0.818 & 0.224 & 3.38\,h \\ %
WilLaGS~\cite{willags} & 25.79 & 0.855 & 0.207 & 27.6\,min \\ %
\midrule
\ours & \textbf{26.00} & \textbf{0.856} & \textbf{0.152} & \textbf{9.7\,min} \\ %
\quad std over seeds & 0.04 & 0.0004 & 0.0004 & 0.08\,min \\
\bottomrule
\end{tabular}
\caption{NeRF-OSR, mean over the four scenes.
Per-scene results are in \extref{tab:supp_osr}.
Bold: best per column.}
\label{tab:osr}
\end{table}

\begin{figure*}[t]
\centering
\includegraphics[width=\textwidth]{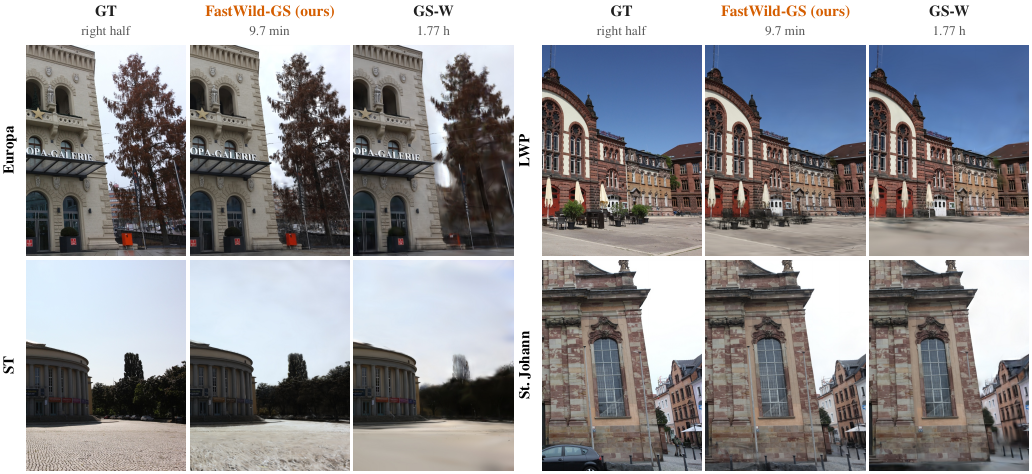}
\caption{Right halves of NeRF-OSR test views rendered by \ours and GS-W.}
\label{fig:osr}
\end{figure*}

\ours reaches 26.00\,dB PSNR, 0.856 SSIM and 0.152 LPIPS on average, the best mean values of all methods, and trains in 9.7 minutes per scene (\cref{tab:osr}).
\ours preserves the trees of Europa and ST, which GS-W blurs (\cref{fig:osr}).

\begin{figure*}[t]
\centering
\includegraphics[width=\textwidth]{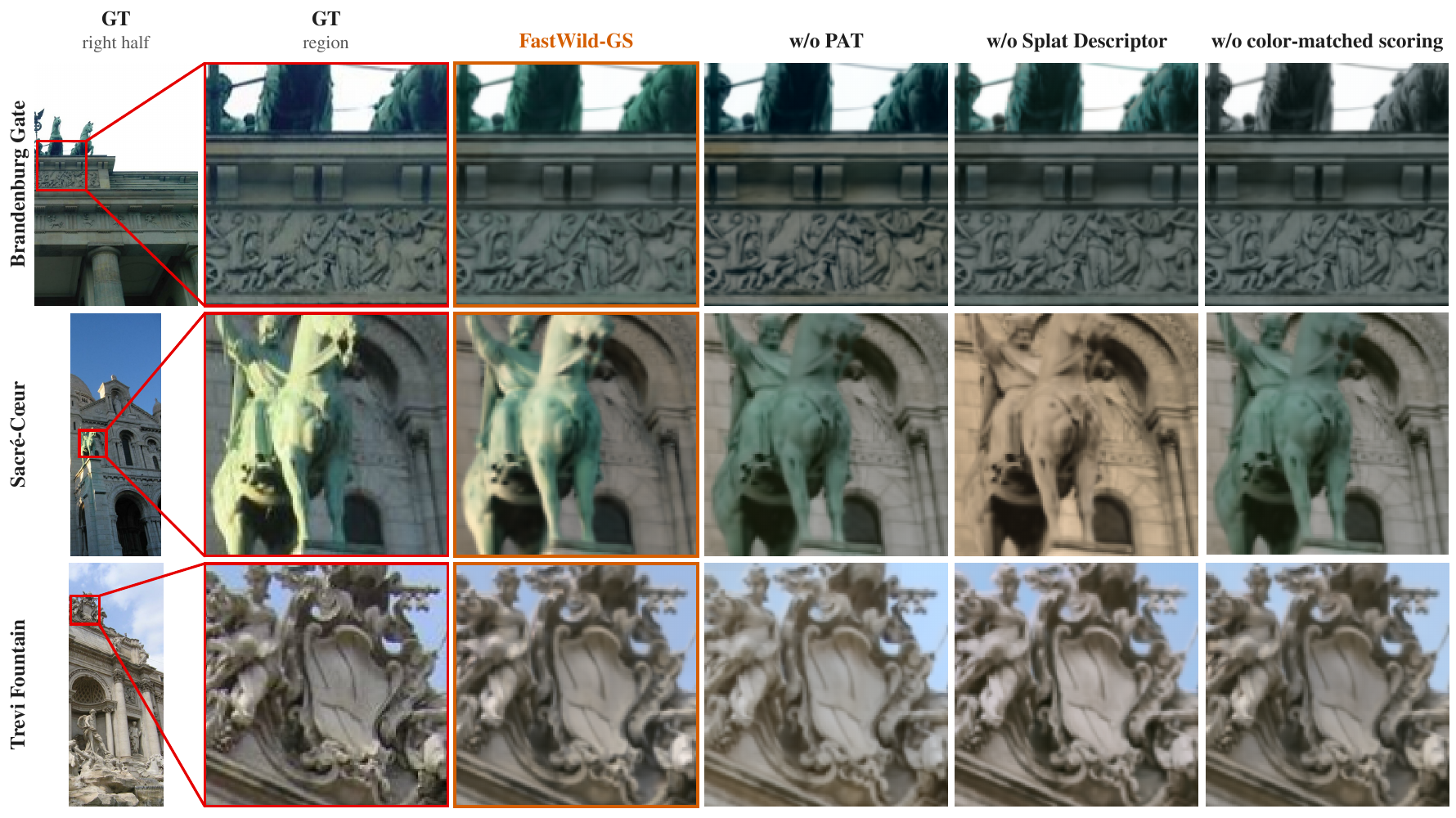}
\caption{Enlarged regions of PhotoTourism test views rendered by \ours and by the variants without the PAT, without the Splat Descriptor and without color-matched scoring.
The first column shows the right half of each test view, with the enlarged region marked in red.}
\label{fig:ablation}
\end{figure*}

\newpage
\null
\newpage
\subsection{Ablations}
\label{sec:ablation}

\begin{table}[t]
\centering
\small
\setlength{\tabcolsep}{3pt} %
\begin{tabular}{@{}l ccc r@{}}
\toprule
Setting & PSNR$\uparrow$ & SSIM$\uparrow$ & LPIPS$\downarrow$ & Train \\
\midrule
\ours & 25.52 & 0.874 & 0.1461 & 8.8\,min \\
\midrule
w/o PAT & 21.22 & 0.834 & 0.1882 & 7.6\,min \\
w/o PAT Decoder & 21.18 & 0.835 & 0.1867 & 8.0\,min \\
w/o Splat Descriptor & 23.53 & 0.861 & 0.1660 & 11.4\,min \\
WildGaussians appearance & 23.54 & 0.856 & 0.1721 & 12.4\,min \\
WilLaGS appearance & 23.09 & 0.853 & 0.1750 & 15.2\,min \\
SWAG appearance & 22.91 & 0.852 & 0.1813 & 10.4\,min \\
\midrule
w/o color-matched scoring & 24.80 & 0.873 & 0.1491 & 8.9\,min \\
w/o channel transform & 24.94 & 0.874 & 0.1491 & 8.8\,min \\
decoder-colored scoring & 24.97 & 0.874 & 0.1492 & 8.6\,min \\
\quad + channel transform & 24.93 & 0.874 & 0.1484 & 8.4\,min \\
\midrule
w/o $\mathcal{L}_{\text{depth}}$ & 25.06 & 0.874 & 0.1488 & 7.0\,min \\
w/o growth schedule & 24.94 & 0.873 & 0.1507 & 8.9\,min \\
\midrule
+ person/vehicle mask & 25.16 & 0.875 & 0.1462 & 9.0\,min \\
\bottomrule
\end{tabular}
\caption{Ablations on PhotoTourism; each row removes, replaces or adds a part of \ours (definitions: \extref{sec:supp_impl}; per-scene results: \extref{tab:supp_ablation_scene}).
Train: training time per scene.}
\label{tab:ablation}
\end{table}

Removing any component of \ours lowers PSNR and raises LPIPS, and adding a person and vehicle mask to the photometric losses lowers PSNR by 0.36\,dB (\cref{tab:ablation}).

\paragraph{Appearance module.}
The largest drop occurs when the color of a Gaussian no longer depends on the image.
Without the PAT or without the PAT Decoder, PSNR falls by 4.30 and 4.34\,dB, respectively (\cref{tab:ablation,fig:ablation}).
The two variants perform almost equally, because in both of them only the channel transform adapts the render to each image.
When the decoder still receives the PAT but not the Splat Descriptor, PSNR falls by 1.99\,dB.
This gap suggests that the learned per-Gaussian feature lets the decoder turn a single PAT into a different color change for each Gaussian.
Replacing our appearance module, including the channel transform, with that of WildGaussians~\cite{kulhanek2024wildgaussians}, which corrects a color stored in each Gaussian, lowers PSNR by 1.98\,dB and lengthens the training from \cOursMin\ to 12.4 minutes.
With the appearance modules of WilLaGS~\cite{willags} and SWAG~\cite{dahmani2024swag}, PSNR falls by 2.43 and 2.61\,dB, respectively.

\paragraph{Color-matched scoring.}
Applying the channel transform to the scoring renders raises PSNR by 0.72\,dB and lowers LPIPS from 0.1491 to 0.1461 without additional renders; the number of Gaussians and the training time are unchanged (\extref{tab:supp_ablation_scene}).
The transform raises PSNR only together with color-matched scoring, as in training alone it gives 24.80\,dB against 24.94\,dB without any transform.
A scoring color that does not depend on the learned appearance is better, and coloring the scoring renders with the PAT Decoder lowers PSNR by 0.55\,dB, and by 0.59\,dB with the channel transform.

\paragraph{Depth loss and growth schedule.}
Without the depth loss, whose prior does not depend on the appearance of the image, PSNR falls by 0.46\,dB.
This variant trains in 7.0 minutes and reaches the LPIPS of AsymGS, 0.150.
The growth schedule raises PSNR by 0.58\,dB without increasing the training time, and ends with more Gaussians (\cref{fig:growth}).

\paragraph{Test-time fitting.}
The radial gain and the channel transform both contribute at test time, and PSNR falls by 0.32\,dB without the former and by 0.71\,dB without the latter (\cref{tab:tta_suppl}).

\subsection{Other Backbones}
\label{sec:backbones}

Attached to six backbones~\cite{huang2024gs2d,yu2024mipsplatting,kheradmand2024mcmc,mallick2024taming,radl2024stopthepop,lu2024scaffold} without any change to their densification or losses, the appearance module of \ours brings all of them to a PSNR between 24.62 and 25.14\,dB (\cref{tab:backbones}).

\begin{table}[t]
\centering
\footnotesize
\setlength{\tabcolsep}{2pt}
\begin{tabular}{@{}l c ccc@{}}
\toprule
Test-time fitting & \#param. & PSNR$\uparrow$ & SSIM$\uparrow$ & LPIPS$\downarrow$ \\
\midrule
\ours: PAT, $s,b$, $v$; 512 steps & 137 & 25.52 & 0.874 & 0.1461 \\ %
\midrule
w/o radial gain $v$ & 134 & 25.20 & 0.874 & 0.1467 \\
w/o channel transform $s,b$ & 131 & 24.81 & 0.872 & 0.1504 \\
128 steps & 137 & 25.13 & 0.874 & 0.1474 \\
256 steps & 137 & 25.45 & 0.874 & 0.1464 \\
1024 steps & 137 & 25.51 & 0.875 & 0.1461 \\
\bottomrule
\end{tabular}
\caption{Test-time fitting on PhotoTourism with the trained models of \ours fixed.}
\label{tab:tta_suppl}
\end{table}

\begin{table}[t]
\centering
\small
\setlength{\tabcolsep}{4pt}
\begin{tabular}{@{}l ccc@{}}
\toprule
Backbone & PSNR$\uparrow$ & SSIM$\uparrow$ & LPIPS$\downarrow$ \\
\midrule
2DGS~\cite{huang2024gs2d} & 17.65 & 0.784 & 0.2618 \\
\quad + Ours & 24.68 & 0.866 & 0.1604 \\
\midrule
Mip-Splatting~\cite{yu2024mipsplatting} & 16.94 & 0.748 & 0.3258 \\
\quad + Ours & 24.62 & 0.867 & 0.1540 \\
\midrule
3DGS-MCMC~\cite{kheradmand2024mcmc} & 18.40 & 0.810 & 0.2168 \\
\quad + Ours & 24.76 & 0.869 & 0.1591 \\
\midrule
Taming-3DGS~\cite{mallick2024taming} & 18.33 & 0.801 & 0.2235 \\
\quad + Ours & 24.72 & 0.868 & 0.1539 \\
\midrule
StopThePop~\cite{radl2024stopthepop} & 18.42 & 0.794 & 0.2382 \\
\quad + Ours & 24.91 & 0.868 & 0.1529 \\
\midrule
Scaffold-GS~\cite{lu2024scaffold} & 18.58 & 0.811 & 0.2053 \\
\quad + Ours & 25.14 & 0.869 & 0.1554 \\
\bottomrule
\end{tabular}
\caption{The appearance module of \ours attached to other backbones on PhotoTourism (mean of the three scenes).
Backbones without it have no appearance model.}
\label{tab:backbones}
\end{table}
\section{Conclusion}
\label{sec:conclusion}

\ours brings in-the-wild Gaussian Splatting to minutes-scale training.
By adapting both the colors and the density control of a shared scene to each image, it trains faster than all compared methods with the rendering quality of methods that train for hours, and its appearance module also extends six other 3DGS backbones to photo collections.
The PAT describes each image with a single vector, and letting the token vary across the image is a natural next step toward local effects, such as a shadow crossing a facade.
\section*{LLM Usage Statement}

We used LLM-based coding agents (Claude, Anthropic; Codex, OpenAI) in the research and in the writing.
In the research, the agents helped refine the research ideas and the experiment design, assisted in writing the training, evaluation and timing code and in computing the metrics, and wrote the code that produces the tables, plots and diagrams.
In the writing, they surveyed related work, edited text written by the authors and gave reviewer-style feedback.
The photographs and rendered images in the figures are dataset images and outputs of the trained models, only cropped and arranged.
The authors take full responsibility for all content.
{
    \small
    \bibliographystyle{ieeenat_fullname}
    \bibliography{main}
}

\onecolumn
\raggedbottom
\makeatletter
\let\@addtocurcol\flafter@addtocurcol
\setlength{\@dblfptop}{0pt}
\setlength{\@dblfpsep}{\dblfloatsep}
\setlength{\@dblfpbot}{0pt plus 1fil}
\makeatother
\setcounter{section}{0}
\setcounter{table}{0}
\setcounter{figure}{0}
\setcounter{equation}{0}
\setcounter{footnote}{0}
\renewcommand{\thesection}{\Alph{section}}
\renewcommand{\thetable}{S\arabic{table}}
\renewcommand{\thefigure}{S\arabic{figure}}
\renewcommand{\theHsection}{supp.\Alph{section}}
\renewcommand{\theHtable}{supp.\arabic{table}}
\renewcommand{\theHfigure}{supp.\arabic{figure}}
\renewcommand{\maketitlesupplementary}{%
    \begin{center}
        {\Large\bfseries \thetitle\par}
        \vspace{0.5em}
        {\large Supplementary Material\par}
        \vspace{1em}
    \end{center}
}
\clearpage
\maketitlesupplementary

\section{Implementation details}
\label{sec:supp_impl}

\Cref{tab:supp_hparams} lists the hyperparameters of training and of test-time fitting.
On PhotoTourism, K-Planes, GS-W, Scaffold-GS and gsplat are trained with the configurations of NerfBaselines~\cite{kulhanek2024nerfbaselines}; all other methods with public code are trained with the official code and default settings.
\begin{table*}[t]
\centering
\small
\setlength{\tabcolsep}{4pt}
\renewcommand{\arraystretch}{1.12}
\newcommand{\hpgroup}[1]{\multicolumn{2}{@{}l}{\rule{0pt}{2.8ex}\textbf{#1}}}
\newcommand{\hpgroupfirst}[1]{\multicolumn{2}{@{}l}{\textbf{#1}}}
\begin{tabular}{@{}p{0.47\textwidth}@{\hspace{0.035\textwidth}}p{0.495\textwidth}@{}}
\toprule
\begin{tabular}[t]{@{}ll@{}}
\hpgroupfirst{Input} \\
Images & original resolution \\
Poses, initial points & public SfM reconstructions \\
\hpgroup{Appearance module} \\
PAT and feature sizes & $e_i \in \mathbb{R}^{128}$, $f_g \in \mathbb{R}^{32}$ \\
PAT Decoder & two hidden layers of width 128, ReLU \\
View direction & without positional encoding \\
Initialization & PAT and $f_g$ at zero, $(s_i,b_i)$ at identity \\
\hpgroup{Optimization} \\
Iterations & 30k \\
Optimizer & Adam \\
Depth weight $\lambda_{\text{d}}(t)$ & $0$ until 0.5k, linear to $0.05$ at 2k, \\
 & constant until 15k, linear to $0.005$ at 30k \\
Parameter averaging & 100 snapshots every 20 iterations \\
 & from 28k, after the last pruning at 27k \\
\hpgroup{NeRF-OSR evaluation} \\
Metric region & right half within the NeRF-OSR masks \\
Training and fitting & without masks \\
Test-time fitting & left half of each test image \\
\end{tabular}
&
\begin{tabular}[t]{@{}ll@{}}
\hpgroupfirst{Learning rates} \\
Position & $1.6{\times}10^{-4}$ to $1.6{\times}10^{-6}$ ($\times$ scene extent) \\
Opacity, scale, rotation & $0.025$, $0.005$, $0.001$ \\
Feature $f_g$ & $0.0025$ \\
PAT, PAT Decoder & $5{\times}10^{-4}$, $10^{-3}$, constant \\
Channel transform & $10^{-3}$ to $10^{-4}$, exponential, 5k warm-up \\
\hpgroup{Density control} \\
Densification & every 100 iterations from 0.6k to 14.9k \\
Gradient thresholds & view-space $0.0002$, absolute $0.0012$ \\
Pixel-error threshold & $0.1$ \\
Threshold scaling & $\times 0.651$ from 12k \\
Additional pruning & at 5.9k, 8.9k, 11.9k and 14.9k \\
Scoring renders & 10 training views per step \\
\hpgroup{Test-time fitting} \\
Optimization & 512 Adam steps, constant learning rates \\
Initialization & all parameters at zero \\
Learning rates & $e^\ast$ $0.02$, $(s^\ast,b^\ast)$ $0.005$, $v$ $0.01$ \\
Photometric loss & $0.8\,\mathcal{L}_{\text{photo}} + 0.2\,(1-\mathrm{SSIM})$ \\
Squared-norm penalty & $10^{-4}$ on $e^\ast$, $s^\ast$, $b^\ast$; $10^{-2}$ on $v$ \\
\end{tabular}
\\
\bottomrule
\end{tabular}
\caption{Hyperparameters of \ours, shared by all scenes of both datasets. Iteration counts are given in thousands (k).}
\label{tab:supp_hparams}
\end{table*}

\begin{table}[t]
\centering
\footnotesize
\begin{tabular}{@{}lccc@{}}
\toprule
Scene & Training time (min) & Gaussians (M) & Training memory (GiB) \\
\midrule
Brandenburg Gate & 6.1 & 0.30 & 2.64 \\ %
Sacr\'e-C\oe ur & 8.5 & 0.45 & 4.28 \\ %
Trevi Fountain & 11.8 & 0.68 & 5.43 \\ %
\midrule
Mean & 8.8 & 0.48 & 4.12 \\ %
\bottomrule
\end{tabular}
\caption{Training time (including the extraction of the depth prior), number of Gaussians and peak GPU memory in training of \ours per scene on PhotoTourism.}
\label{tab:supp_time}
\end{table}

\paragraph{Ablation variants.}
\emph{w/o PAT}: the PAT Decoder takes $[f_g, d_{g,i}]$ only, and test-time fitting fits only the channel transform and the radial gain.
\emph{w/o PAT Decoder}: each Gaussian is colored by trained spherical-harmonics coefficients, without the PAT.
\emph{w/o Splat Descriptor}: the PAT Decoder takes a trained spherical-harmonics color of the Gaussian in direction $d_{g,i}$ instead of $[f_g, d_{g,i}]$.
\emph{WildGaussians appearance}: the appearance module is replaced by that of WildGaussians~\cite{kulhanek2024wildgaussians}, in which an MLP predicts an affine transform of a color stored in each Gaussian from a per-image embedding of size 32 and a per-Gaussian embedding of size 24; there is no channel transform, the scoring renders use the fixed SfM color, and test-time fitting fits the embedding of the test image in the same 512 steps.
\emph{WildGaussians appearance + channel transform, color-matched scoring}: as WildGaussians appearance, with the channel transform of \ours in training and in the scoring renders, which use the fixed SfM color, and test-time fitting fits the embedding together with $(s^\ast,b^\ast)$ and $v$ in the same 512 steps.
\emph{WilLaGS appearance}: the appearance module is replaced by that of WilLaGS~\cite{willags}, following its description, in which a $\beta$-VAE trained from scratch ($\beta=2$, loss weight 0.01) encodes each image into a latent code of size 64, a tri-plane field of three $64\times64$ planes with 32 channels gives a feature of size 32 at the position of each Gaussian, and a three-layer MLP predicts the color from this feature, a per-Gaussian feature of size 32 and the view direction; there is no channel transform, the scoring renders use the colors that this module predicts from the mean latent code of each training image, and test-time fitting starts from the latent code of the left half of the test image and fits it in the same 512 steps.
\emph{SWAG appearance}: the appearance module is replaced by that of SWAG~\cite{dahmani2024swag}, following its description, in which an MLP predicts the color from hash-grid features at the position of each Gaussian, a per-image embedding of size 24 and the spherical-harmonics color, without the transient opacity of SWAG; there is no channel transform, the scoring renders use the colors that this module predicts from the embedding of each training image, and test-time fitting fits the embedding of the test image in the same 512 steps.
\emph{w/o color-matched scoring}: the scoring renders do not use the channel transform.
\emph{w/o channel transform}: no $(s_i,b_i)$ in training or in the scoring renders; test-time fitting still fits $(s^\ast,b^\ast)$.
\emph{decoder-colored scoring}: the scoring renders are colored by the PAT Decoder instead of the fixed SfM color, without and with the channel transform.
\emph{w/o growth schedule}: the densification thresholds of FastGS throughout and no additional pruning steps.
\emph{w/o growth schedule, 0.48\,M Gaussians}: no additional pruning steps, and the densification thresholds of FastGS are scaled at all densification steps by a factor per scene, between 0.82 and 0.84, that brings the final number of Gaussians to that of \ours.
\emph{w/o parameter averaging}: the parameters at 30k iterations are used without averaging.
\emph{w/o RMS term}: $\mathcal{L}_{\text{photo}}$ is the L1 term alone in training.
\emph{+ person/vehicle mask}: pixels that Mask2Former~\cite{cheng2022mask2former} labels as person, bicycle, car, motorcycle, bus or truck are excluded from the photometric losses during training; the depth loss, the scoring renders and test-time fitting are unchanged.

\section{Additional results}
\label{sec:supp_results}

\begin{table}[t]
\centering
\footnotesize
\setlength{\tabcolsep}{2pt} %
\begin{tabular}{@{}l ccc ccc r@{}}
\toprule
Setting & PAT & $f_g$ & Fitted & $\Delta$PSNR & $\Delta$SSIM & $\Delta$LPIPS & Train \\
\midrule
\ours & 128 & 32 & 137 & & & & 8.8\,min \\
PAT size 64 & 64 & 32 & 73 & \ensuremath{-}0.26 & \ensuremath{-}0.0014 & +0.0022 & 8.5\,min \\
PAT size 256 & 256 & 32 & 265 & \ensuremath{-}0.30 & +0.0001 & +0.0007 & 9.7\,min \\
feature $f_g$ size 16 & 128 & 16 & 137 & \ensuremath{-}0.06 & \ensuremath{-}0.0002 & +0.0007 & 8.4\,min \\
feature $f_g$ size 64 & 128 & 64 & 137 & \ensuremath{-}0.29 & \ensuremath{-}0.0004 & +0.0015 & 9.5\,min \\
\bottomrule
\end{tabular}
\caption{Size of the Photo Appearance Token (PAT) and of the per-Gaussian feature $f_g$ on PhotoTourism (mean of the three scenes).
$\Delta$: difference to \ours. Fitted: number of parameters fitted per test view.}
\label{tab:supp_size}
\end{table}
\begin{table*}[t]
\centering
\footnotesize
\setlength{\tabcolsep}{5pt} %
\resizebox{\textwidth}{!}{%
\begin{tabular}{@{}l ccc ccc ccc ccc r r@{}}
\toprule
 & \multicolumn{3}{c}{Brandenburg Gate} & \multicolumn{3}{c}{Sacr\'e-C\oe ur} & \multicolumn{3}{c}{Trevi Fountain} & \multicolumn{3}{c}{Mean} & & \\
\cmidrule(lr){2-4}\cmidrule(lr){5-7}\cmidrule(lr){8-10}\cmidrule(lr){11-13}
Setting & PSNR$\uparrow$ & SSIM$\uparrow$ & LPIPS$\downarrow$ & PSNR$\uparrow$ & SSIM$\uparrow$ & LPIPS$\downarrow$ & PSNR$\uparrow$ & SSIM$\uparrow$ & LPIPS$\downarrow$ & PSNR$\uparrow$ & SSIM$\uparrow$ & LPIPS$\downarrow$ & Gaussians & Train \\
\midrule
\ours & 28.87 & 0.940 & 0.1086 & 23.72 & 0.893 & 0.1357 & 23.96 & 0.790 & 0.1941 & 25.52 & 0.874 & 0.1461 & 0.48\,M & 8.8\,min \\
\midrule
w/o PAT & 25.02 & 0.918 & 0.1341 & 18.36 & 0.846 & 0.1815 & 20.29 & 0.738 & 0.2491 & 21.22 & 0.834 & 0.1882 & 0.45\,M & 7.6\,min \\
w/o PAT Decoder & 25.02 & 0.919 & 0.1337 & 18.31 & 0.847 & 0.1820 & 20.21 & 0.738 & 0.2443 & 21.18 & 0.835 & 0.1867 & 0.46\,M & 8.0\,min \\
w/o Splat Descriptor & 27.29 & 0.938 & 0.1162 & 21.11 & 0.871 & 0.1605 & 22.21 & 0.773 & 0.2212 & 23.53 & 0.861 & 0.1660 & 0.42\,M & 11.4\,min \\
WildGaussians appearance & 27.42 & 0.933 & 0.1228 & 20.78 & 0.864 & 0.1685 & 22.41 & 0.771 & 0.2249 & 23.54 & 0.856 & 0.1721 & 0.55\,M & 12.4\,min \\
\quad + channel transform, color-matched scoring & 27.32 & 0.936 & 0.1158 & 21.90 & 0.882 & 0.1482 & 22.85 & 0.779 & 0.2107 & 24.02 & 0.866 & 0.1582 & 0.46\,M & 13.5\,min \\
WilLaGS appearance & 25.65 & 0.919 & 0.1417 & 20.95 & 0.868 & 0.1624 & 22.68 & 0.772 & 0.2210 & 23.09 & 0.853 & 0.1750 & 0.42\,M & 15.2\,min \\
SWAG appearance & 26.07 & 0.924 & 0.1383 & 21.38 & 0.869 & 0.1679 & 21.28 & 0.762 & 0.2377 & 22.91 & 0.852 & 0.1813 & 0.39\,M & 10.4\,min \\
\midrule
w/o color-matched scoring & 27.57 & 0.940 & 0.1116 & 23.01 & 0.890 & 0.1393 & 23.84 & 0.790 & 0.1963 & 24.80 & 0.873 & 0.1491 & 0.48\,M & 8.9\,min \\
w/o channel transform & 27.87 & 0.940 & 0.1116 & 22.93 & 0.890 & 0.1410 & 24.02 & 0.791 & 0.1945 & 24.94 & 0.874 & 0.1491 & 0.48\,M & 8.8\,min \\
\midrule
w/o $\mathcal{L}_{\text{depth}}$ & 28.32 & 0.940 & 0.1102 & 22.86 & 0.891 & 0.1404 & 23.99 & 0.790 & 0.1960 & 25.06 & 0.874 & 0.1488 & 0.48\,M & 7.0\,min \\
w/o growth schedule & 28.06 & 0.940 & 0.1122 & 22.84 & 0.890 & 0.1421 & 23.93 & 0.789 & 0.1978 & 24.94 & 0.873 & 0.1507 & 0.40\,M & 8.9\,min \\
\quad 0.48\,M Gaussians & 28.07 & 0.940 & 0.1102 & 23.29 & 0.892 & 0.1379 & 23.87 & 0.790 & 0.1965 & 25.08 & 0.874 & 0.1482 & 0.48\,M & 10.3\,min \\
w/o parameter averaging & 27.76 & 0.939 & 0.1138 & 22.99 & 0.888 & 0.1415 & 23.77 & 0.785 & 0.1978 & 24.84 & 0.871 & 0.1510 & 0.48\,M & 8.8\,min \\
w/o RMS term & 28.29 & 0.940 & 0.1110 & 22.42 & 0.890 & 0.1410 & 23.90 & 0.790 & 0.1949 & 24.87 & 0.873 & 0.1490 & 0.47\,M & 8.8\,min \\
\midrule
decoder-colored scoring & 27.70 & 0.939 & 0.1125 & 23.22 & 0.891 & 0.1392 & 23.99 & 0.792 & 0.1959 & 24.97 & 0.874 & 0.1492 & 0.46\,M & 8.6\,min \\
\quad + channel transform & 27.83 & 0.940 & 0.1113 & 23.01 & 0.891 & 0.1393 & 23.94 & 0.791 & 0.1948 & 24.93 & 0.874 & 0.1484 & 0.45\,M & 8.4\,min \\
\midrule
+ person/vehicle mask & 28.60 & 0.943 & 0.1039 & 22.92 & 0.891 & 0.1387 & 23.96 & 0.791 & 0.1959 & 25.16 & 0.875 & 0.1462 & 0.47\,M & 9.0\,min \\
\bottomrule
\end{tabular}%
}
\caption{Per-scene results on PhotoTourism of the ablations in \extref{tab:ablation} and of the further variants of \cref{sec:supp_impl}, with their means over the three scenes.
Gaussians: number of Gaussians at the end of training, mean over the three scenes.
The training time of + person/vehicle mask excludes the mask extraction.}
\label{tab:supp_ablation_scene}
\end{table*}
\begin{table*}[t]
\centering
\footnotesize
\setlength{\tabcolsep}{3pt}
\resizebox{\textwidth}{!}{%
\begin{tabular}{@{}l ccc ccc ccc ccc ccc@{}}
\toprule
 & \multicolumn{3}{c}{Europa} & \multicolumn{3}{c}{LWP} & \multicolumn{3}{c}{ST} & \multicolumn{3}{c}{St.\ Johann} & \multicolumn{3}{c}{Mean} \\
\cmidrule(lr){2-4}\cmidrule(lr){5-7}\cmidrule(lr){8-10}\cmidrule(lr){11-13}\cmidrule(lr){14-16}
Method & PSNR$\uparrow$ & SSIM$\uparrow$ & LPIPS$\downarrow$ & PSNR$\uparrow$ & SSIM$\uparrow$ & LPIPS$\downarrow$ & PSNR$\uparrow$ & SSIM$\uparrow$ & LPIPS$\downarrow$ & PSNR$\uparrow$ & SSIM$\uparrow$ & LPIPS$\downarrow$ & PSNR$\uparrow$ & SSIM$\uparrow$ & LPIPS$\downarrow$ \\
\midrule
NeRF-MS~\cite{li2023nerfms} & 21.03 & 0.721 & 0.294 & 21.90 & 0.719 & 0.336 & 20.68 & 0.630 & 0.402 & 22.84 & 0.793 & 0.235 & 21.61 & 0.716 & 0.317 \\
SWAG~\cite{dahmani2024swag} & 23.74 & 0.845 & 0.242 & 22.07 & 0.783 & 0.303 & 22.29 & 0.713 & 0.364 & 23.91 & 0.864 & 0.172 & 23.00 & 0.801 & 0.270 \\
GS-W~\cite{zhang2024gsw} & 22.04 & 0.758 & 0.231 & 24.44 & 0.827 & 0.208 & 22.45 & 0.690 & 0.315 & 26.23 & 0.893 & 0.113 & 23.79 & 0.792 & 0.217 \\
WE-GS~\cite{wang2024wegs} & 24.73 & \textbf{0.873} & 0.157 & 24.33 & 0.821 & 0.197 & 22.45 & 0.720 & 0.341 & 24.12 & 0.858 & 0.202 & 23.91 & 0.818 & 0.224 \\
WilLaGS~\cite{willags} & \textbf{25.50} & 0.830 & 0.230 & \textbf{27.07} & \textbf{0.879} & 0.200 & \textbf{24.45} & 0.788 & 0.255 & 26.13 & 0.925 & 0.144 & 25.79 & 0.855 & 0.207 \\
\midrule
\ours & 23.79 & 0.842 & \textbf{0.155} & 26.00 & 0.845 & \textbf{0.178} & 23.89 & \textbf{0.792} & \textbf{0.209} & \textbf{30.31} & \textbf{0.946} & \textbf{0.067} & \textbf{26.00} & \textbf{0.856} & \textbf{0.152} \\ %
\quad std over seeds & 0.03 & 0.0008 & 0.0020 & 0.01 & 0.0007 & 0.0005 & 0.06 & 0.0007 & 0.0008 & 0.15 & 0.0015 & 0.0017 & 0.04 & 0.0004 & 0.0004 \\
\bottomrule
\end{tabular}%
}
\caption{NeRF-OSR per scene.
Bold: best per column.}
\label{tab:supp_osr}
\end{table*}

\begin{figure*}[t]
\centering
\includegraphics[width=\textwidth]{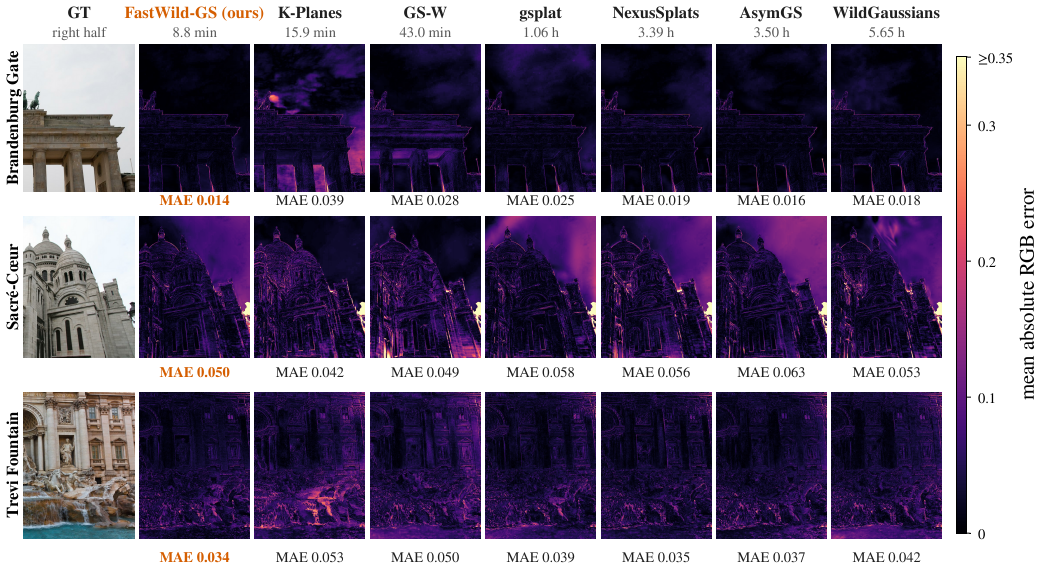}
\caption{Per-pixel error on the right halves of the test views of \extref{fig:teaser}: mean absolute RGB error against the ground truth.
In each scene, the view is the landscape test view of median LPIPS for \ours.
Below each map: the mean absolute error of the shown half.}
\label{fig:supp_errmap}
\end{figure*}

\Cref{tab:supp_time} gives the training time, the number of Gaussians and the training memory per scene, \cref{tab:supp_size} the results for other sizes of the PAT and of the per-Gaussian feature, \cref{tab:supp_ablation_scene,tab:supp_osr} the per-scene values of \extref{tab:ablation} and \extref{tab:osr}, \cref{tab:supp_std,tab:supp_std_bb,tab:supp_std_scene} the standard deviations over seeds, and \cref{fig:supp_errmap} the per-pixel error on the views of \extref{fig:teaser}.

\paragraph{Transient objects.}
Excluding the pixels that Mask2Former labels as persons or vehicles from the photometric losses lowers the mean PSNR on PhotoTourism by 0.36\,dB (\cref{tab:supp_ablation_scene}).

\section{Further analysis}
\label{sec:supp_analysis}

\subsection{Evaluation sources on PhotoTourism}
\label{sec:supp_sources}
The values of K-Planes, GS-W, Scaffold-GS, gsplat and WildGaussians in \extref{tab:main} follow the half-image protocol of \mbox{NerfBaselines}~\cite{kulhanek2024nerfbaselines}, which fits the appearance embedding of each method on the left half of a test view and evaluates the right half, as our protocol does.
AsymGS~\cite{asymgs} gives its values in its own paper, which does not state a half-image protocol, and WilLaGS~\cite{willags} evaluates AsymGS and WildGaussians under the half-image protocol (\cref{tab:supp_pt_sources}).
For WildGaussians, the means of this evaluation agree with those of \extref{tab:main} within 0.14\,dB in PSNR, 0.001 in SSIM and 0.002 in LPIPS.
In this evaluation, AsymGS reaches 25.03\,dB PSNR, 0.850 SSIM and 0.186 LPIPS, against 25.62\,dB, 0.872 and 0.150 for its GS-GS variant in its own paper (\extref{tab:main}).
The training time of AsymGS in \extref{tab:main} is that of this variant.
\ours is better than the half-image evaluation of AsymGS in all three metrics on every scene, by 0.49\,dB in PSNR, 0.025 in SSIM and 0.040 in LPIPS on average.
Image-conditioned methods such as GS-W~\cite{zhang2024gsw}, Wild-GS~\cite{xu2024wildgs} and WE-GS~\cite{wang2024wegs} obtain the appearance of a test view with an image encoder, and their papers do not state a half-image protocol; Wild-GS also evaluates at half resolution.
When the encoder sees the full test view, it also sees the evaluated right half, and NerfBaselines gives GS-W 19.73\,dB on Sacr\'e-C\oe ur under the half-image protocol, against 23.24\,dB in the GS-W paper.
We therefore list GS-W from NerfBaselines in \extref{tab:main} and no values of image-conditioned methods that come without the half-image protocol.
For training time, WilLaGS gives 0.9 GPU hours per scene on an RTX~3090 GPU in its paper, and our reproduction of WilLaGS trains in 30.8 minutes on the RTX~5090 GPU of all our measurements.

\begin{table}[H]
\centering
\footnotesize
\setlength{\tabcolsep}{3pt}
\begin{tabular}{@{}l ccc ccc ccc ccc@{}}
\toprule
 & \multicolumn{3}{c}{Brandenburg Gate} & \multicolumn{3}{c}{Sacr\'e-C\oe ur} & \multicolumn{3}{c}{Trevi Fountain} & \multicolumn{3}{c}{Mean} \\
\cmidrule(lr){2-4}\cmidrule(lr){5-7}\cmidrule(lr){8-10}\cmidrule(lr){11-13}
Method (evaluation) & PSNR$\uparrow$ & SSIM$\uparrow$ & LPIPS$\downarrow$ & PSNR$\uparrow$ & SSIM$\uparrow$ & LPIPS$\downarrow$ & PSNR$\uparrow$ & SSIM$\uparrow$ & LPIPS$\downarrow$ & PSNR$\uparrow$ & SSIM$\uparrow$ & LPIPS$\downarrow$ \\
\midrule
WildGaussians (\extref{tab:main}) & 27.77 & 0.927 & 0.133 & 22.56 & 0.859 & 0.177 & 23.63 & 0.766 & 0.228 & 24.65 & 0.851 & 0.179 \\
WildGaussians (WilLaGS) & 27.46 & 0.927 & 0.141 & 22.61 & 0.856 & 0.173 & 23.47 & 0.768 & 0.228 & 24.51 & 0.850 & 0.181 \\
\midrule
AsymGS (AsymGS paper) & 28.56 & 0.938 & 0.109 & 23.78 & 0.887 & 0.139 & 24.52 & 0.790 & 0.202 & 25.62 & 0.872 & 0.150 \\
AsymGS (WilLaGS) & 28.19 & 0.924 & 0.146 & 23.29 & 0.862 & 0.169 & 23.61 & 0.763 & 0.243 & 25.03 & 0.850 & 0.186 \\
\midrule
WilLaGS (WilLaGS) & 28.45 & 0.937 & 0.134 & 23.80 & 0.889 & 0.150 & 24.27 & 0.822 & 0.168 & 25.51 & 0.883 & 0.151 \\
\midrule
\ours & 28.87 & 0.940 & 0.1086 & 23.72 & 0.893 & 0.1357 & 23.96 & 0.790 & 0.1941 & 25.52 & 0.874 & 0.1461 \\
\quad w/o radial gain $v$ & 28.48 & 0.940 & 0.1089 & 23.35 & 0.892 & 0.1369 & 23.77 & 0.789 & 0.1945 & 25.20 & 0.874 & 0.1467 \\
\quad 128 fitting steps & 28.01 & 0.940 & 0.1102 & 23.42 & 0.892 & 0.1382 & 23.97 & 0.790 & 0.1938 & 25.13 & 0.874 & 0.1474 \\
\quad PAT only & 27.37 & 0.937 & 0.1132 & 23.09 & 0.889 & 0.1394 & 23.73 & 0.787 & 0.1969 & 24.73 & 0.871 & 0.1498 \\
\quad PAT only, 128 fitting steps & 27.17 & 0.936 & 0.1167 & 23.06 & 0.888 & 0.1407 & 23.73 & 0.785 & 0.1980 & 24.65 & 0.870 & 0.1518 \\
\bottomrule
\end{tabular}
\caption{PhotoTourism results of WildGaussians, AsymGS and WilLaGS from two evaluation sources.
The WildGaussians values of \extref{tab:main} follow the half-image protocol of NerfBaselines~\cite{kulhanek2024nerfbaselines}, and WilLaGS~\cite{willags} fits the appearance parameters on the left half of each test view and evaluates the right half.
Below \ours, the first two settings are those of \extref{tab:tta_suppl}, and PAT only fits only the PAT at test time.}
\label{tab:supp_pt_sources}
\end{table}
\begin{table}[t]
\centering
\footnotesize
\setlength{\tabcolsep}{4pt}
\begin{tabular}{@{}l ccc@{}}
\toprule
Setting & PSNR & SSIM & LPIPS \\
\midrule
\multicolumn{4}{@{}l}{\emph{Ablations (\extref{tab:ablation} and \cref{tab:supp_ablation_scene})}} \\
\ours & 0.15 & 0.0003 & 0.0017 \\
w/o PAT & 0.02 & 0.0000 & 0.0010 \\
w/o PAT Decoder & 0.01 & 0.0002 & 0.0012 \\
w/o Splat Descriptor & 0.11 & 0.0011 & 0.0007 \\
WildGaussians appearance & 0.06 & 0.0006 & 0.0009 \\
\quad + channel transform, color-matched scoring & 0.05 & 0.0011 & 0.0009 \\
WilLaGS appearance & 0.09 & 0.0013 & 0.0018 \\
SWAG appearance & 0.03 & 0.0013 & 0.0016 \\
w/o color-matched scoring & 0.12 & 0.0005 & 0.0012 \\
w/o channel transform & 0.07 & 0.0004 & 0.0005 \\
w/o $\mathcal{L}_{\text{depth}}$ & 0.07 & 0.0006 & 0.0010 \\
w/o growth schedule & 0.18 & 0.0010 & 0.0014 \\
\quad 0.48\,M Gaussians & 0.05 & 0.0005 & 0.0006 \\
w/o parameter averaging & 0.12 & 0.0006 & 0.0015 \\
w/o RMS term & 0.07 & 0.0009 & 0.0009 \\
decoder-colored scoring & 0.07 & 0.0003 & 0.0013 \\
\quad + channel transform & 0.06 & 0.0002 & 0.0007 \\
+ person/vehicle mask & 0.02 & 0.0004 & 0.0008 \\
\midrule
\multicolumn{4}{@{}l}{\emph{Test-time fitting (\extref{tab:tta_suppl} and \cref{tab:supp_pt_sources})}} \\
w/o radial gain $v$ & 0.15 & 0.0004 & 0.0016 \\
w/o channel transform $s,b$ & 0.08 & 0.0007 & 0.0006 \\
128 steps & 0.04 & 0.0002 & 0.0016 \\
256 steps & 0.13 & 0.0002 & 0.0018 \\
1024 steps & 0.17 & 0.0003 & 0.0018 \\
PAT only & 0.24 & 0.0006 & 0.0013 \\
PAT only, 128 steps & 0.17 & 0.0005 & 0.0012 \\
\midrule
\multicolumn{4}{@{}l}{\emph{Sizes (\cref{tab:supp_size}), of the difference to \ours}} \\
PAT size 64 & 0.01 & 0.0000 & 0.0015 \\
PAT size 256 & 0.18 & 0.0004 & 0.0023 \\
feature $f_g$ size 16 & 0.05 & 0.0015 & 0.0017 \\
feature $f_g$ size 64 & 0.16 & 0.0001 & 0.0012 \\
\bottomrule
\end{tabular}
\caption{Standard deviation over seeds of the PhotoTourism means in \extref{tab:ablation}, \cref{tab:supp_ablation_scene}, \extref{tab:tta_suppl}, \cref{tab:supp_size} and \cref{tab:supp_pt_sources}.}
\label{tab:supp_std}
\end{table}
\begin{table}[t]
\centering
\footnotesize
\setlength{\tabcolsep}{4pt}
\begin{tabular}{@{}l ccc@{}}
\toprule
Backbone & PSNR & SSIM & LPIPS \\
\midrule
2DGS & 0.09 & 0.0086 & 0.0150 \\
\quad + embedding & 0.14 & 0.0022 & 0.0038 \\
\quad + Ours & 0.02 & 0.0009 & 0.0016 \\
\midrule
Mip-Splatting & 0.16 & 0.0091 & 0.0134 \\
\quad + embedding & 0.13 & 0.0013 & 0.0022 \\
\quad + Ours & 0.01 & 0.0021 & 0.0014 \\
\midrule
3DGS-MCMC & 0.09 & 0.0015 & 0.0013 \\
\quad + embedding & 0.07 & 0.0007 & 0.0009 \\
\quad + Ours & 0.06 & 0.0006 & 0.0015 \\
\midrule
Taming-3DGS & 0.03 & 0.0009 & 0.0035 \\
\quad + embedding & 0.08 & 0.0012 & 0.0013 \\
\quad + Ours & 0.06 & 0.0010 & 0.0012 \\
\midrule
StopThePop & 0.20 & 0.0043 & 0.0080 \\
\quad + embedding & 0.07 & 0.0017 & 0.0013 \\
\quad + Ours & 0.04 & 0.0003 & 0.0008 \\
\midrule
Scaffold-GS & 0.10 & 0.0017 & 0.0036 \\
\quad + embedding & 0.10 & 0.0008 & 0.0016 \\
\quad + Ours & 0.13 & 0.0002 & 0.0009 \\
\bottomrule
\end{tabular}
\caption{Standard deviation over seeds of the values in \extref{tab:backbones} and \cref{tab:supp_backbones_emb}.}
\label{tab:supp_std_bb}
\end{table}
\begin{table*}[t]
\centering
\footnotesize
\setlength{\tabcolsep}{4pt}
\begin{tabular}{@{}l ccc ccc ccc@{}}
\toprule
 & \multicolumn{3}{c}{Brandenburg Gate} & \multicolumn{3}{c}{Sacr\'e-C\oe ur} & \multicolumn{3}{c}{Trevi Fountain} \\
\cmidrule(lr){2-4}\cmidrule(lr){5-7}\cmidrule(lr){8-10}
Setting & PSNR & SSIM & LPIPS & PSNR & SSIM & LPIPS & PSNR & SSIM & LPIPS \\
\midrule
\ours & 0.39 & 0.0001 & 0.0029 & 0.20 & 0.0009 & 0.0017 & 0.04 & 0.0010 & 0.0021 \\
w/o PAT & 0.01 & 0.0000 & 0.0004 & 0.05 & 0.0004 & 0.0021 & 0.01 & 0.0004 & 0.0006 \\
w/o PAT Decoder & 0.00 & 0.0000 & 0.0003 & 0.00 & 0.0002 & 0.0008 & 0.04 & 0.0010 & 0.0032 \\
w/o Splat Descriptor & 0.36 & 0.0006 & 0.0013 & 0.01 & 0.0040 & 0.0031 & 0.02 & 0.0002 & 0.0002 \\
WildGaussians appearance & 0.07 & 0.0003 & 0.0000 & 0.13 & 0.0011 & 0.0019 & 0.04 & 0.0013 & 0.0020 \\
\quad + channel transform, color-matched scoring & 0.14 & 0.0006 & 0.0008 & 0.10 & 0.0020 & 0.0023 & 0.06 & 0.0026 & 0.0020 \\
WilLaGS appearance & 0.21 & 0.0024 & 0.0027 & 0.15 & 0.0016 & 0.0018 & 0.06 & 0.0007 & 0.0018 \\
SWAG appearance & 0.13 & 0.0029 & 0.0022 & 0.10 & 0.0019 & 0.0021 & 0.05 & 0.0015 & 0.0018 \\
w/o color-matched scoring & 0.31 & 0.0012 & 0.0030 & 0.12 & 0.0015 & 0.0010 & 0.04 & 0.0008 & 0.0011 \\
w/o channel transform & 0.05 & 0.0003 & 0.0010 & 0.18 & 0.0009 & 0.0020 & 0.05 & 0.0007 & 0.0004 \\
w/o $\mathcal{L}_{\text{depth}}$ & 0.17 & 0.0003 & 0.0006 & 0.12 & 0.0010 & 0.0021 & 0.06 & 0.0005 & 0.0005 \\
w/o growth schedule & 0.28 & 0.0006 & 0.0006 & 0.22 & 0.0016 & 0.0023 & 0.08 & 0.0011 & 0.0014 \\
\quad 0.48\,M Gaussians & 0.21 & 0.0006 & 0.0008 & 0.16 & 0.0012 & 0.0024 & 0.05 & 0.0012 & 0.0007 \\
w/o parameter averaging & 0.32 & 0.0007 & 0.0027 & 0.10 & 0.0010 & 0.0014 & 0.02 & 0.0014 & 0.0014 \\
w/o RMS term & 0.15 & 0.0008 & 0.0014 & 0.11 & 0.0009 & 0.0013 & 0.06 & 0.0017 & 0.0010 \\
decoder-colored scoring & 0.15 & 0.0008 & 0.0022 & 0.15 & 0.0012 & 0.0021 & 0.05 & 0.0009 & 0.0008 \\
\quad + channel transform & 0.25 & 0.0003 & 0.0018 & 0.20 & 0.0007 & 0.0015 & 0.04 & 0.0003 & 0.0013 \\
+ person/vehicle mask & 0.10 & 0.0004 & 0.0016 & 0.04 & 0.0010 & 0.0018 & 0.02 & 0.0002 & 0.0011 \\
\bottomrule
\end{tabular}
\caption{Standard deviation over seeds of the per-scene values in \cref{tab:supp_ablation_scene}.}
\label{tab:supp_std_scene}
\end{table*}

\subsection{Test-time fitting}
\label{sec:supp_tta}
At test time, \ours fits the PAT, the channel transform and the radial gain of \extref{eq:radial} on the left half of each view, and the radial gain enters only test-time fitting.
Without the radial gain, \ours reaches 25.20\,dB PSNR, 0.874 SSIM and 0.1467 LPIPS, and with 128 instead of 512 fitting steps it reaches 25.13\,dB, 0.874 and 0.1474 (\extref{tab:tta_suppl}).
In \cref{tab:supp_pt_sources}, these two settings lie above the half-image evaluation of AsymGS in all three metrics, and their SSIM and LPIPS differ by at most 0.0013 from those of the full fitting.
With only the PAT fitted, in 512 or 128 steps, \ours has a higher SSIM and a lower LPIPS than the half-image evaluation of AsymGS on every scene.

\subsection{Structural quality across variants}
\label{sec:supp_structure}
SSIM compares local structure and contrast, LPIPS compares deep features, and both depend less on the global color of a render than PSNR.
The 19 settings of \extref{tab:ablation}, \extref{tab:tta_suppl}, \cref{tab:supp_size} and \cref{tab:supp_pt_sources} that keep the PAT, the PAT Decoder and the Splat Descriptor include the variants without the depth loss, without the radial gain, without color-matched scoring, with 128 fitting steps, with a PAT of half the size and with only the PAT fitted.
Across these settings, SSIM lies between 0.870 and 0.875 and LPIPS between 0.1461 and 0.1518.
The whole range lies above the half-image evaluation of AsymGS in SSIM and below it in LPIPS (0.850 and 0.186).

\subsection{Results after color correction}
\label{sec:supp_colorcorr}
After a per-channel scale and bias is fitted to each render by least squares against the ground truth of the evaluated right half, \ours has the highest PSNR and SSIM and the lowest LPIPS of the methods in \cref{tab:supp_colorcorr}, 26.37\,dB, 0.8731 and 0.1449.
\begin{table}[H]
\centering
\footnotesize
\setlength{\tabcolsep}{5pt}
\begin{tabular}{@{}l ccc ccc@{}}
\toprule
 & \multicolumn{3}{c}{Before} & \multicolumn{3}{c}{After color correction} \\
\cmidrule(lr){2-4}\cmidrule(l){5-7}
Method & PSNR$\uparrow$ & SSIM$\uparrow$ & LPIPS$\downarrow$ & PSNR$\uparrow$ & SSIM$\uparrow$ & LPIPS$\downarrow$ \\
\midrule
K-Planes~\cite{fridovichkeil2023kplanes} & 21.10 & 0.7612 & 0.3133 & 22.20 & 0.7605 & 0.3076 \\
GS-W~\cite{zhang2024gsw} & 21.38 & 0.8172 & 0.2126 & 22.91 & 0.8147 & 0.2118 \\
gsplat~\cite{ye2025gsplat} & 23.66 & 0.8573 & 0.1619 & 25.05 & 0.8585 & 0.1587 \\
WildGaussians~\cite{kulhanek2024wildgaussians} & 24.65 & 0.8506 & 0.1792 & 25.63 & 0.8494 & 0.1782 \\
NexusSplats~\cite{nexussplats} & 24.91 & 0.8492 & 0.1856 & 25.73 & 0.8485 & 0.1841 \\
AsymGS~\cite{asymgs} & 25.01 & 0.8521 & 0.1857 & 26.01 & 0.8513 & 0.1846 \\
\midrule
\ours & \textbf{25.52} & \textbf{0.8745} & \textbf{0.1461} & \textbf{26.37} & \textbf{0.8731} & \textbf{0.1449} \\
\bottomrule
\end{tabular}
\caption{PhotoTourism (mean of the three scenes) before and after the color correction of \cref{sec:supp_colorcorr}.
All values are computed with our evaluation code.
Bold: best per column.}
\label{tab:supp_colorcorr}
\end{table}

\subsection{How color-matched scoring acts}
\label{sec:supp_cms}
The scoring renders of \extref{sec:scoring} are computed without gradients, enter no loss and only decide which Gaussians are densified and which are pruned.
\ours and its variant without color-matched scoring share the losses, the appearance module and the test-time fitting with the channel transform and the radial gain, and they differ only in the color of the scoring renders.
The 0.72\,dB in PSNR and the 0.0030 in LPIPS between them (\extref{tab:ablation}) therefore come from the Gaussians that density control keeps.
Both end with 0.48\,M Gaussians and train in 8.8 and 8.9 minutes (\cref{tab:supp_ablation_scene}), and color-matched scoring changes which Gaussians density control keeps at the same number of Gaussians.
The channel transform improves the reconstruction through the same path.
Used in training and in the scoring renders, it raises PSNR from 24.94 to 25.52\,dB.

The choice of the scoring color follows from the same ablations.
Coloring the scoring renders with the PAT Decoder, which is fitted to each scored image, ends training with 0.46\,M instead of 0.48\,M Gaussians, takes 8.6 instead of 8.8 minutes and gives 0.55\,dB lower PSNR (\cref{tab:supp_ablation_scene}).
The fixed SfM color with the channel transform gives the best reconstruction among the scoring colors of \extref{tab:ablation}.
Without color-matched scoring, a pixel-error threshold of 0.05, 0.15 or 0.2 instead of 0.1 gives 24.73 to 24.97\,dB, a least-squares color fit of each scoring render to its image gives 25.09\,dB, and the PAT Decoder color with a zero PAT gives 25.01\,dB, against 25.52\,dB with color-matched scoring (\cref{tab:supp_cms_controls}).
\begin{table}[H]
\centering
\footnotesize
\setlength{\tabcolsep}{5pt}
\begin{tabular}{@{}l ccc r@{}}
\toprule
Scoring renders & PSNR$\uparrow$ & SSIM$\uparrow$ & LPIPS$\downarrow$ & Gaussians \\
\midrule
SfM color with the channel transform (\ours) & 25.52 & 0.874 & 0.1461 & 0.48\,M \\
\midrule
SfM color, pixel-error threshold 0.05 & 24.73 & 0.874 & 0.1488 & 0.48\,M \\
SfM color, pixel-error threshold 0.1 (w/o color-matched scoring) & 24.80 & 0.873 & 0.1491 & 0.48\,M \\
SfM color, pixel-error threshold 0.15 & 24.97 & 0.874 & 0.1491 & 0.47\,M \\
SfM color, pixel-error threshold 0.2 & 24.89 & 0.873 & 0.1493 & 0.45\,M \\
SfM color with a least-squares color fit to the image & 25.09 & 0.874 & 0.1482 & 0.48\,M \\
PAT Decoder color with a zero PAT & 25.01 & 0.874 & 0.1487 & 0.47\,M \\
\bottomrule
\end{tabular}
\caption{Scoring renders without color-matched scoring on PhotoTourism (mean of the three scenes).
The pixel-error threshold of \ours is 0.1 (\cref{tab:supp_hparams}).
The least-squares color fit applies to each scoring render the per-channel scale and bias that best match it to its image.
Gaussians: number of Gaussians at the end of training.}
\label{tab:supp_cms_controls}
\end{table}

\subsection{Roles of the parts of the appearance module}
\label{sec:supp_roles}
Without the channel transform, the appearance module of \ours reaches 24.94\,dB PSNR, 0.874 SSIM and 0.1491 LPIPS, against 23.54\,dB, 0.856 and 0.1721 for that of WildGaussians, and our module trains in 29\% less time, 8.8 instead of 12.4 minutes (\extref{tab:ablation}).
The channel transform, applied in training and in the scoring renders, then raises PSNR to 25.52\,dB and lowers LPIPS to 0.1461.
With the same channel transform, the appearance module of WildGaussians reaches 24.02\,dB, 0.866 SSIM and 0.1582 LPIPS (\cref{tab:supp_ablation_scene}).

\subsection{The appearance module on other backbones}
\label{sec:supp_backbones}
NerfBaselines adds per-image appearance embeddings to Scaffold-GS~\cite{lu2024scaffold} and gsplat~\cite{ye2025gsplat}, which reach 0.854 and 0.857 SSIM and 0.170 and 0.162 LPIPS in \extref{tab:main}.
In \extref{tab:backbones}, + Ours adds the Splat Descriptor, the PAT, the PAT Decoder and the channel transform to each backbone, and test-time fitting fits the PAT, $(s^\ast,b^\ast)$ and $v$ as in \ours.
With the appearance module of \ours, each of the six backbones of \extref{tab:backbones} reaches a higher SSIM, between 0.866 and 0.869, and a lower LPIPS, between 0.1529 and 0.1604.
With a per-image embedding instead, the six backbones reach 22.64 to 22.99\,dB PSNR, against 24.62 to 25.14\,dB with the appearance module of \ours (\cref{tab:supp_backbones_emb}).
\begin{table}[H]
\centering
\footnotesize
\setlength{\tabcolsep}{5pt}
\begin{tabular}{@{}l ccc r@{}}
\toprule
Backbone & PSNR$\uparrow$ & SSIM$\uparrow$ & LPIPS$\downarrow$ & Train \\
\midrule
2DGS~\cite{huang2024gs2d} & 17.65 & 0.784 & 0.2618 & 12.8\,min \\
\quad + embedding & 22.66 & 0.843 & 0.1864 & 23.5\,min \\
\quad + Ours & 24.68 & 0.866 & 0.1604 & 20.0\,min \\
\midrule
Mip-Splatting~\cite{yu2024mipsplatting} & 16.94 & 0.748 & 0.3258 & 11.2\,min \\
\quad + embedding & 22.78 & 0.847 & 0.1767 & 23.7\,min \\
\quad + Ours & 24.62 & 0.867 & 0.1540 & 19.1\,min \\
\midrule
3DGS-MCMC~\cite{kheradmand2024mcmc} & 18.40 & 0.810 & 0.2168 & 11.2\,min \\
\quad + embedding & 22.87 & 0.849 & 0.1826 & 20.1\,min \\
\quad + Ours & 24.76 & 0.869 & 0.1591 & 15.7\,min \\
\midrule
Taming-3DGS~\cite{mallick2024taming} & 18.33 & 0.801 & 0.2235 & 4.0\,min \\
\quad + embedding & 22.99 & 0.851 & 0.1743 & 15.3\,min \\
\quad + Ours & 24.72 & 0.868 & 0.1539 & 11.4\,min \\
\midrule
StopThePop~\cite{radl2024stopthepop} & 18.42 & 0.794 & 0.2382 & 8.3\,min \\
\quad + embedding & 22.83 & 0.847 & 0.1761 & 20.2\,min \\
\quad + Ours & 24.91 & 0.868 & 0.1529 & 15.8\,min \\
\midrule
Scaffold-GS~\cite{lu2024scaffold} & 18.58 & 0.811 & 0.2053 & 10.8\,min \\
\quad + embedding & 22.64 & 0.847 & 0.1777 & 13.7\,min \\
\quad + Ours & 25.14 & 0.869 & 0.1554 & 15.7\,min \\
\bottomrule
\end{tabular}
\caption{Other backbones on PhotoTourism (mean of the three scenes) without an appearance model, with a per-image embedding and with the appearance module of \ours (\extref{tab:backbones}).
+ embedding: a per-image embedding of size 32 and an MLP that adds a color offset to the color of the backbone; its test-time fitting fits the embedding of the test image in 512 steps.
Train: training time per scene.}
\label{tab:supp_backbones_emb}
\end{table}

\subsection{The depth loss}
\label{sec:supp_depth}
Like the SfM color of the scoring renders, the depth prior from Depth Anything 3 does not depend on the appearance of an image.
Without the depth loss, \ours trains in 7.0 minutes and reaches 25.06\,dB, 0.874 SSIM and 0.1488 LPIPS (\extref{tab:ablation}).
It matches the SSIM and LPIPS that AsymGS gives in its own paper (0.872 and 0.150) and is better than the half-image evaluation of AsymGS in both (0.850 and 0.186).

\subsection{Parameter averaging, the RMS term and the growth schedule}
\label{sec:supp_training}
Without parameter averaging, \ours reaches 24.84\,dB, 0.871 SSIM and 0.1510 LPIPS, and without the RMS term of $\mathcal{L}_{\text{photo}}$, 24.87\,dB, 0.873 and 0.1490 (\cref{tab:supp_ablation_scene}).
Without the growth schedule, densification thresholds lowered to end at the same 0.48\,M Gaussians give 25.08\,dB, 0.874 and 0.1482 in 10.3 minutes, against 25.52\,dB, 0.874 and 0.1461 in 8.8 minutes with the growth schedule.

\subsection{NeRF-OSR with a common evaluation of the compared methods}
\label{sec:supp_osr_sources}
WilLaGS also evaluates GS-W and WildGaussians on the four NeRF-OSR scenes under the half-image protocol (\cref{tab:supp_osr_sources}).
With the compared methods taken from this evaluation, \ours has the highest mean PSNR and SSIM, and it exceeds GS-W and WildGaussians in PSNR and SSIM on every scene.

\begin{table}[H]
\centering
\footnotesize
\setlength{\tabcolsep}{3pt}
\resizebox{\textwidth}{!}{%
\begin{tabular}{@{}l ccc ccc ccc ccc ccc@{}}
\toprule
 & \multicolumn{3}{c}{Europa} & \multicolumn{3}{c}{LWP} & \multicolumn{3}{c}{ST} & \multicolumn{3}{c}{St.\ Johann} & \multicolumn{3}{c}{Mean} \\
\cmidrule(lr){2-4}\cmidrule(lr){5-7}\cmidrule(lr){8-10}\cmidrule(lr){11-13}\cmidrule(lr){14-16}
Method & PSNR$\uparrow$ & SSIM$\uparrow$ & LPIPS$\downarrow$ & PSNR$\uparrow$ & SSIM$\uparrow$ & LPIPS$\downarrow$ & PSNR$\uparrow$ & SSIM$\uparrow$ & LPIPS$\downarrow$ & PSNR$\uparrow$ & SSIM$\uparrow$ & LPIPS$\downarrow$ & PSNR$\uparrow$ & SSIM$\uparrow$ & LPIPS$\downarrow$ \\
\midrule
WildGaussians~\cite{kulhanek2024wildgaussians} & 20.82 & 0.724 & 0.393 & 22.12 & 0.791 & 0.289 & 19.90 & 0.695 & 0.375 & 22.77 & 0.881 & 0.184 & 21.40 & 0.773 & 0.310 \\
GS-W~\cite{zhang2024gsw} & 22.42 & 0.792 & 0.379 & 23.66 & 0.818 & 0.334 & 22.91 & 0.750 & 0.406 & 27.19 & 0.898 & 0.238 & 24.04 & 0.815 & 0.339 \\
WilLaGS~\cite{willags} & 25.50 & 0.830 & 0.230 & 27.07 & 0.879 & 0.200 & 24.45 & 0.788 & 0.255 & 26.13 & 0.925 & 0.144 & 25.79 & 0.855 & 0.207 \\
\midrule
\ours & 23.79 & 0.842 & 0.155 & 26.00 & 0.845 & 0.178 & 23.89 & 0.792 & 0.209 & 30.31 & 0.946 & 0.067 & 26.00 & 0.856 & 0.152 \\
\bottomrule
\end{tabular}%
}
\caption{NeRF-OSR per scene with the compared methods taken from the half-image evaluation of WilLaGS~\cite{willags}.
The values of \ours are those of \cref{tab:supp_osr}.}
\label{tab:supp_osr_sources}
\end{table}

\subsection{Rendering and fitting cost}
\label{sec:supp_cost}
The PAT Decoder is an MLP with an input of 163 values, $[f_g, d_{g,i}, e_i]$, two hidden layers of width 128 and three outputs, which amounts to 37{,}632 multiply-accumulate operations per Gaussian and view.
The part of the first layer that acts on the PAT is the same for all Gaussians of a view and is computed once per view, which leaves 21{,}248 operations per Gaussian.
For the 0.30 to 0.68\,M Gaussians of the PhotoTourism scenes (\cref{tab:supp_time}), the decoder needs 6.4 to 14.4\,G multiply-accumulate operations per view.
At the original resolution, \ours renders a test view at 244 to 511 frames per second with at most 1.39\,GiB of GPU memory, stores 114 to 255\,MB per scene, and fits a test view in 2.6 to 4.9 seconds (\cref{tab:supp_efficiency}).
\begin{table}[H]
\centering
\footnotesize
\setlength{\tabcolsep}{5pt}
\begin{tabular}{@{}l cccc@{}}
\toprule
Scene & Rendering (FPS) & Rendering memory (GiB) & Model size (MB) & Fitting (s per view) \\
\midrule
Brandenburg Gate & 511 & 0.63 & 114 & 2.64 \\
Sacr\'e-C\oe ur & 349 & 0.94 & 170 & 3.60 \\
Trevi Fountain & 244 & 1.39 & 255 & 4.86 \\
\midrule
Mean & 368 & 0.99 & 180 & 3.70 \\
\bottomrule
\end{tabular}
\caption{Rendering and test-time fitting cost of \ours per scene on PhotoTourism at the original resolution.
Rendering memory: peak GPU memory in rendering.
Fitting: test-time fitting time per test view.}
\label{tab:supp_efficiency}
\end{table}

\end{document}